\documentclass[preprint,12pt]{elsarticle}

\usepackage[export]{adjustbox}
\usepackage{lscape}
\usepackage{rotating}

\usepackage{float}

\usepackage{graphicx}

\usepackage{amssymb}

\usepackage[table]{xcolor}

\usepackage{lineno}
\usepackage{multirow}
\usepackage{subcaption}
\usepackage{array}
\usepackage{colortbl}
\usepackage{hhline}
\usepackage{amsmath} 
\usepackage{tabularx}
\usepackage{booktabs}

\usepackage{hyperref}
\usepackage{longtable}

\usepackage{acronym}

\acrodef{FNAC}{Fine-needle Aspiration Cytology}
\acrodef{CNN}{Convolutional Neural Network}
\acrodef{ANN}{Artificial Neural Network}
\acrodef{SVM}{Support Vector Machine}
\acrodef{kNN}{K-Nearest Neighbors}
\acrodef{FFT}{Fast Fourier Transform}
\acrodef{DL}{Deep Learning}
\acrodef{GAN}{Generative Adversarial Networks}
\acrodef{SLR}{Systematic Literature Review}
\acrodef{RF}{Random Forest}
\acrodef{NCA}{Neighborhood Component Analysis}
\acrodef{RVM}{Relevance Vector Machine}
\acrodef{DBSCAN}{Density-based spatial clustering of applications with noise}
\acrodef{SLIC}{Simple Linear Iterative Clustering}
\acrodef{ML}{Machine Learning}
\acrodef{TVT}{Train-Validation-Test}
\acrodef{LDA}{Linear Discriminant Analysis}
\acrodef{SSAE}{Stacked Sparse Autoencoder}
\acrodef{FCRN}{Fully Convolutional Regression Network}
\acrodef{NORs}{Nucleolar Organizer Regions}
\acrodef{HOG}{Histogram of Oriented Gradients}
\acrodef{PCA}{Principal Component Analysis}
\acrodef{MIL}{Multiple Instance Learning}

\journal{CMIG Computerized Medical Imaging and Graphics}

\begin{document}

\begin{frontmatter}

\title{What is the State of the Art of Computer Vision-Assisted Cytology? \\A Systematic Literature Review}

\author[ine]{André Victória Matias \corref{cor}}
\ead{andre.v.matias@posgrad.ufsc.br}
\author[ine]{João Gustavo Atkinson Amorim}
\ead{joao.atkinson@posgrad.ufsc.br}
\author[auto]{Luiz Antonio Buschetto Macarini}
\ead{luiz.buschetto@posgrad.ufsc.br}
\author[ine]{Allan Cerentini}
\ead{allan.c@posgrad.ufsc.br}

\author[clinic]{Alexandre Sherlley Casimiro Onofre}
\ead{alexandre.onofre@ufsc.br}
\author[clinic]{Fabiana Botelho De Miranda Onofre}
\ead{fabiana.onofre@ufsc.br}
\author[pathology]{Felipe Perozzo Daltoé}
\ead{felipe.daltoe@ufsc.br}

\author[auto]{Marcelo Ricardo Stemmer}
\ead{marcelo.stemmer@ufsc.br}
\author[incod]{Aldo von Wangenheim}
\ead{aldo.vw@ufsc.br}

\address[ine]{Department of Informatics and Statistics, Federal University of Santa Catarina, Florianópolis, Brazil}
\address[auto]{Automation and Systems Department, Federal University of Santa Catarina, Florianópolis, Brazil}
\address[pathology]{Department
of Pathology, Federal University of Santa Catarina, Florianópolis, Brazil}
\address[clinic]{Clinical Analyses Department, Federal University of Santa Catarina, Florianópolis, Brazil}
\address[incod]{Brazilian Institute for Digital Convergence, Federal University of Santa Catarina, Florianópolis, Brazil}

\cortext[cor]{Corresponding author}

\begin{abstract}

Cytology is a low-cost and non-invasive diagnostic procedure employed to support the diagnosis of a broad range of pathologies. Cells are harvested from tissues by aspiration or scraping, and it is still predominantly performed manually by medical or laboratory professionals extensively trained for this purpose. It is a time-consuming and repetitive process where many diagnostic criteria are subjective and vulnerable to human interpretation. Computer Vision technologies, by automatically generating quantitative and objective descriptions of examinations' contents, can help minimize the chances of misdiagnoses and shorten the time required for analysis. To identify the state-of-art of computer vision techniques currently applied to cytology, we conducted a Systematic Literature Review, searching for approaches for the segmentation, detection, quantification, and classification of cells and organelles using computer vision on cytology slides. We analyzed papers published in the last 4 years. The initial search was executed in September 2020 and resulted in 431 articles. After applying the inclusion/exclusion criteria, 157 papers remained, which we analyzed to build a picture of the tendencies and problems present in this research area, highlighting the computer vision methods, staining techniques, evaluation metrics, and the availability of the used datasets and computer code. As a result, we identified that the most used methods in the analyzed works are deep learning-based (70 papers), while fewer works employ classic computer vision only (101 papers). The most recurrent metric used for classification and object detection was the accuracy (33 papers and 5 papers), while for segmentation it was the Dice Similarity Coefficient (38 papers). Regarding staining techniques, Papanicolaou was the most employed one (130 papers), followed by H\&E (20 papers) and Feulgen (5 papers). Twelve of the datasets used in the papers are publicly available, with the DTU/Herlev dataset being the most used one. We conclude that there still is a lack of high-quality datasets for many types of stains and most of the works are not mature enough to be applied in a daily clinical diagnostic routine. We also identified a growing tendency towards adopting deep learning-based approaches as the methods of choice.

\end{abstract}


\end{frontmatter}

\section{Introduction}
\label{sec:intro}

    Cytology is a diagnostic technique where cells are harvested from tissues by aspiration or scraping, prepared with different staining techniques and examined on the microscope.  Since first described by Dr. Alfred François Donné in 1837 when he discovered \emph{Trichomonas vaginalis}, this technique has been used to support the diagnosis of a broad range of pathologies. It became popular in 1941 when Dr. Georgios Papanicolaou proved cytology was useful in the diagnostic of malignant cells of vaginal smears of the uterus \cite{Diamantis2013}.

    Cytology has been successfully applied to the diagnostic of some of the most common and deadly cancers of the human body such as those from bone marrow, lungs, pancreas, breast, cervix, thyroid, oral cavity, and many others. Cytology has also been useful to diagnose a complex variety of non-cancer related pathologies such as inflammatory conditions and also pathogens as fungus, bacteria, and virus \cite{Ivanovic2013}. Collecting cells by aspiration or scraping methods presents the advantage of having a lower cost and being less invasive and easier to be performed by clinicians when compared to other methods of harvesting tissue samples for diagnosis, such as the tissue biopsy \cite{Bedrossian2007}.
    
    In cytology, the pathological status of the cells that have been collected can be analyzed by different approaches, such as DNA quantification, identification of cell surface key proteins, presence of pathogens, and, most often, cell morphology, describing the shape and size of cells and their organelles. Each method requires different cell staining and/or processing to highlight specific organelles or parts of the cell, but regardless of the method, the final analysis is usually performed manually under a light microscope \cite{Ivanovic2013}. Although this procedure is performed by medical professionals extensively trained for that, many of the diagnostic criteria are therefore vulnerable to human interpretation and bias \cite{Chapman2011}. Hence, more recent diagnostic approaches are associated with different diagnostic methods and computer technologies in order to reduce the chances of misdiagnoses. Another reason to employ computational methods is the long processing time that a manual analysis process affords \cite{Li2017}. Computer-based methods have the potential of not only increasing the degree of objectiveness and reproducibility of analyses but also their speed \cite{Mehrotra2011}.

    As the algorithms become more robust and the image analysis techniques more powerful, more information from cytology samples can be collected in an automated way \cite{Kandemir2015}. 

    Machine Learning, especially Artificial Neural Networks (ANNs), have been employed, with varying results, to develop Computer Vision (CV) methods for the automated support of quantitative analysis and diagnostic of cytological samples for over 25 years now, with methods ranging from spatial information parameter extraction with graphs and posterior ANN classification \cite{Kolles94, Kolles96}, comparison between ANNs and multivariate statistical methods \cite{Kolles93, Kolles95}, and comparison between simple ANNs, self-organizing semantic maps and especially developed nearest-neighbor-based ANN models \cite{Kolles97}. 

    In the last years, however, ANN-based Deep Learning (DL) methods have revolutionized the field of Neural Networks \cite{krizhevsky2012}. This revolution occurred mainly in image and signal analysis tasks, and achieved remarkable performance in different computational tasks, showing its robustness and effectiveness on feature extraction from data in different scenarios \cite{LeCun2015}. Recent works using DL approaches in genomics and biomedical applications demonstrated the  flexibility of this approach in handling complex problems, turning DL methods into a very promising approach for automatically analysing cytometry data \cite{Li2017}.

\begin{figure}[ht]
    \centering
    \includegraphics[width=1.0\linewidth]{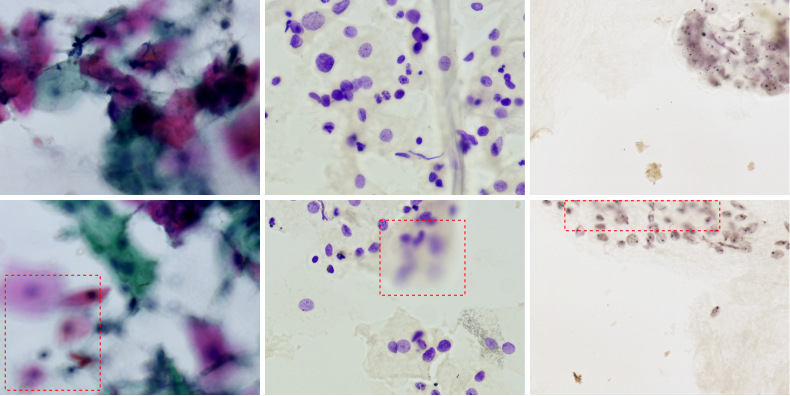}
    \caption{Examples of problems that can be found on the clinical routine. In the first row, highly overlapping cells. On the second row, inside the red-dotted rectangles, unfocused parts of the image fields.}
    \label{fig:intro_figure}
    \end{figure}

    On the other side, several factors can potentially introduce sample acquisition variations, such as microscope particularities, the quality of the staining chemicals, and the lab process. Considering this, to provide reliable cytology CV-based analysis methods, it is necessary to develop algorithms that are robust to this kind of variation. Figure \ref{fig:intro_figure} shows some examples of cytology image fields acquired in a Brazilian laboratory to illustrate such variations.
    
Another scenario that reliable CV methods for cytology could make possible is that of \emph{large-scale tele-cytology}: personnel in labs in small cities in upstate locations are trained to perform the examination collection, samples staining, and image acquisition through digitizing. These samples are then run through image processing routines to perform the quantitative analyses and feature extractions that can be automated. These CV analyses can be performed either on the cloud or local computers, but the software tools will be able to be operated by local upstate personnel. Later, automatically generated reports from these analyses, together with the original digitized slides, can be uploaded into a telemedicine system, where a specialist can review the images and the quantitative data and provide a final findings report. This method would enable a more distributed, much faster, and less specialist-dependent cytology procedure, concentrating the workforce of experts on providing the final analyses and conclusive findings reports. However, to implement such a scenario, it is necessary to develop more robust CV techniques that account for larger quality variations on sample preparation and image acquisition.  
This scenario means that mature, robust, and clinically applicable CV methods in cytology could not only provide better means to perform quantitative cytology but could also represent a paradigm change in a diagnostic process that, even in a digital era, in many places still depends largely on the physical transportation of samples. To inquire into the state-of-the-art CV methods for cytology, we performed a systematic review of the literature (SLR). 

\section{Materials and Methods}

\begin{figure}[H]
    \centering
    \includegraphics[width=1.0\linewidth]{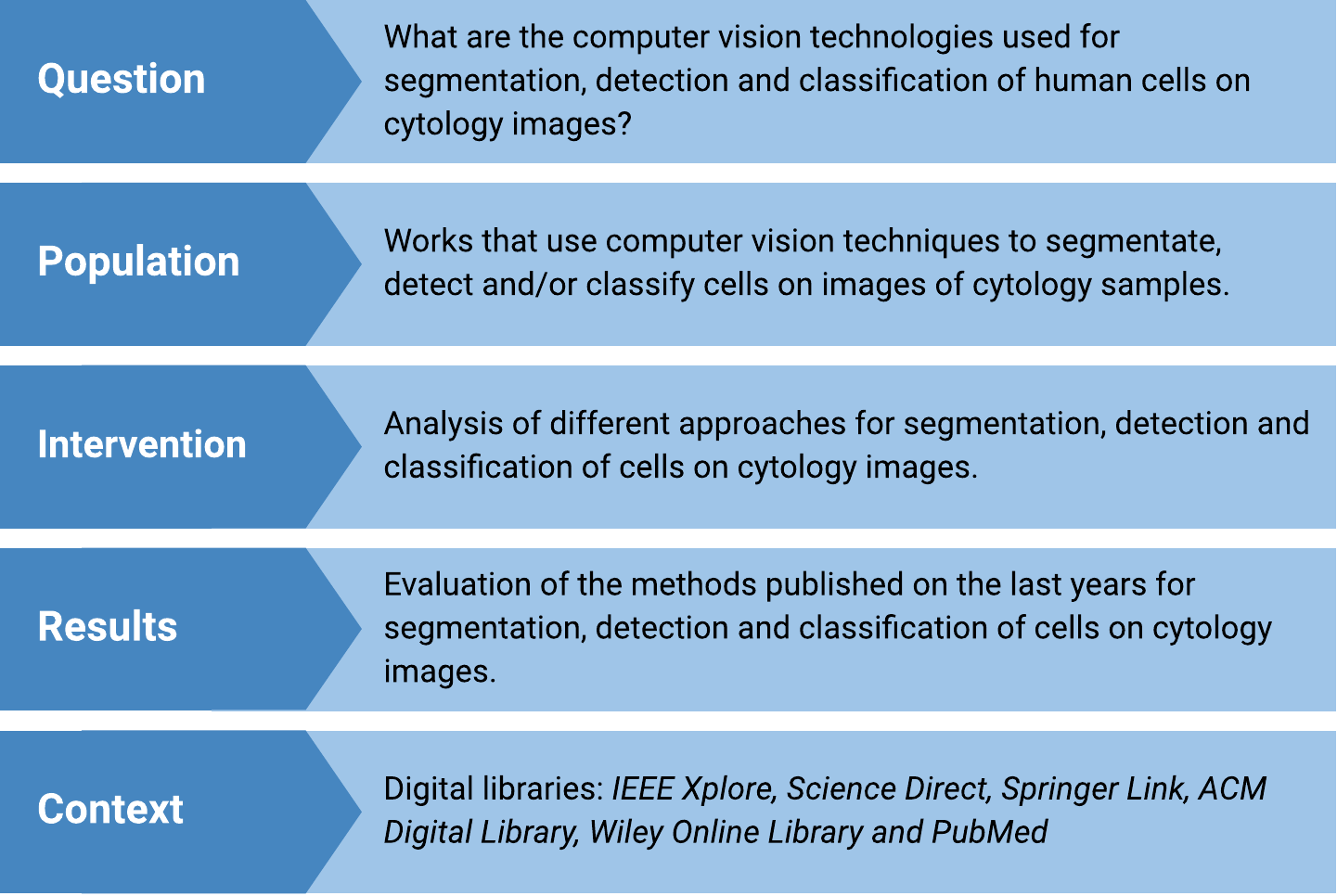}
    \caption{Review definition}
    \label{fig:methodology_figure}
    \end{figure}

    A SLR is conducted based on the systematic review protocols for the Computer Sciences field elaborated by Kitchenham \cite{barbara-slr}. In this review, we searched for approaches for cell segmentation, detection, and classification using computer vision on cytology slides images. Analysing papers published in the last 5 years (from 2016 to 2020), we evaluated the tendencies and main problems present in this research area, highlighting the computer vision methods, staining techniques, result evaluation metrics, and the availability of the used datasets. Our review structure definition is shown in Figure \ref{fig:methodology_figure}. In this review, we focused on identifying commonalities, differences, and tendencies among the different works described in the literature. We provide concise, graphic, and tabular representations of these characteristics and relationships and discuss them.  We also published a complementary technical report \cite{amorim:2020} where we focused on analysing some of the papers individually and in more detail.
    
    \subsection{Search Definitions}
        The details of the search locations (databases), terms, inclusion and exclusion criteria are presented in Tables \ref{tab:ieeexplore_sd}, \ref{tab:science_sd}, \ref{tab:springer_sd}, \ref{tab:acm_sd}, \ref{tab:wiley_sd}, and \ref{tab:pubmed_sd}. The Papanicolaou is the only cytology technique explicitly used in the search terms because a considerable amount of authors does not cite it as a cytology or cytopathology examination and relevant papers would not have been included if these more specific terms were not used.

        To avoid \emph{evidence selection bias}, \cite{Drucker2016} we did not limit our search only to metasearch databases such as \emph{PubMed} or \emph{NCBI}, which tend to present only results from references previously classified as belonging to health sciences and biomedical topics. Instead, we additionally conducted our SLR as it is commonly performed in the field of Computer Sciences (CS), also accessing directly the databases of the individual scientific publishers. We did this because this SLR is focused on \emph{CS research on technology applied to cytology} and a requirement was that relevant CS technology journals and conferences should also be fully taken into consideration. Table \ref{tab:ieeexplore_sd} presents the IEEEXplore Search Definitions.

        \begin{table}[htp!]
        \centering
        \resizebox{\columnwidth}{!}{
            \begin{tabular}{|l|l|}
                \hline
                \multicolumn{1}{|c|}{\textbf{Search Locations}}                                                                                                                                                                                  & \multicolumn{1}{c|}{\textbf{Search Terms}}                                                                                                                                                                                             \\ \hline
                \begin{tabular}[c]{@{}l@{}}-  IEEE Xplore;\\ -  Science Direct;\\ -  Springer Link;\\ -  ACM Digital Library;\\ - Wiley Online Library;\\ - PubMed.\end{tabular}                                                                               & \begin{tabular}[c]{@{}l@{}}- “cytology” OR “cytopathology” OR “Papanicolaou” OR “Pap smear”;\\ - “image” OR “vision”;\\ - “classification” OR “segmentation” OR “detection”;\end{tabular}                                                                               \\ \hline
                \multicolumn{2}{|c|}{\textbf{Inclusion Criteria}}                                                                                                                                                                                                                                                                                                                                                                                                                         \\ \hline
                \multicolumn{2}{|l|}{\begin{tabular}[c]{@{}l@{}}-  Papers that use computer vision to segment, detect and/or classify cells on cytology images;\\ - Studies performed in humans.\end{tabular}}                                                                                                                                                                                                                                        \\ \hline
                \multicolumn{2}{|c|}{\textbf{Exclusion Criteria}}                                                                                                                                                                                                                                                                                                                                                                                                                         \\ \hline
                \multicolumn{2}{|l|}{\begin{tabular}[c]{@{}l@{}}- Papers that do not use computer vision techniques to analyse cytology images;\\ - Studies not performed on human cells;\\ - Papers that focus on image enhancement;\\ - Short papers, search definition, supers, such as abstracts or expanded abstracts;\\ - Review articles;\\ - Articles that were not completely available on the database.\\
                - Papers not written in English.\end{tabular}} \\ \hline
            \end{tabular}
            }
            \caption{Search definition}
            \label{tab:sd}
        \end{table}
    
    \subsection{Search Execution}
    
        The initial search was made on September 16th, 2020 and resulted in 431 papers which were exported to Rayyan QCRI\footnote{https://rayyan.qcri.org/} where 13 duplicates were removed and the remaining ones were divided between four reviewers that read the titles and abstracts of each paper and filtered them based on the inclusion/exclusion criteria. After reading the full texts, the reviewers selected 157 papers in total.

\begin{table}[H]
\resizebox{\columnwidth}{!}{

}
\caption{IEEEXplore Search Definitions}
\label{tab:ieeexplore_sd}
\end{table}

Next, Table \ref{tab:science_sd} presents the Science Direct Search Definitions.

\begin{table}[H]
%
\caption{Science Direct Search Definitions}
\label{tab:science_sd}
\end{table}

In Table \ref{tab:springer_sd} is presented the Springer Link search definitions.

\begin{table}[H]
%
\caption{Springer Link Search Definitions}
\label{tab:springer_sd}
\end{table}

The ACM Digital Library Search Definitions are shown in Table \ref{tab:acm_sd}.

\begin{table}[H]
\resizebox{\columnwidth}{!}{
%
}
\caption{ACM Digital Library Search Definitions}
\label{tab:acm_sd}
\end{table}

In Table \ref{tab:wiley_sd} the Wiley Online Library search definitions are shown.

\begin{table}[H]
\resizebox{\columnwidth}{!}{
%
}
\caption{Wiley Online Search Definitions}
\label{tab:wiley_sd}
\end{table}

Next, in Table \ref{tab:pubmed_sd}, the PubMed Search Definitions are presented.

\begin{table}[H]
%
} \\ \hline
\end{tabular}
\caption{PubMed Search Definitions}
\label{tab:pubmed_sd}
\end{table}

\section{Analysis of the Papers}

In this section, we present the analysis of the papers included in the review. Section \ref{staining-techniques} presents the staining techniques used in the analyzed papers along with a short explanation of each one and Section \ref{datasets} describes the public datasets we found while reading the papers. In Section \ref{image-classification}, we explain the different techniques used for analysis of cytology images in the analyzed papers, grouping them by method (Image Classification, Object Detection, and Segmentation) and by approach (Classic, Deep Learning, and Hybrid).

\subsection{Related Reviews}

Similar works describing systematic reviews on cytology were found during the analysis of the article. Those works were excluded from our SLR for not presenting a method. It is, however, important to cite and analyse them:  As shown in  Table \ref{tab:relatedreviews}, these review papers focus on specific anatomies, specifically \emph{breast} and \emph{cervix}, whereas our SLR addresses Cytology in general. Another important aspect to be noticed is that most of these works focus on specific methods such as ML or DL, whereas we considered both classic CV and machine learning-based methods. Taking this into consideration, we understand that this work has a much larger scope than the related works and also includes the review of more recent articles than most of the cited works.

\begin{table}[]
\begin{tabular}{|c|c|c|c|c|}
\hline
\textbf{Paper}                & \textbf{Year}   & \textbf{Quantity} & \textbf{Body part} & \textbf{Method}  \\ \hline
\cite{WILLIAM201815}          & 2002-2017       & 30                  & breast             & Machine Learning \\ \hline
\cite{surveycytodeeplearning} & until Feb, 2021 & 33                  & cervix             & Deep Learning    \\ \hline
\cite{Sarwar2020}             & 1977 - 2018     & 78                  & cervix             & Segmentation     \\ \hline
\cite{SAHA2016461}            & 1965 - 2016     & 137                 & breast             & Any              \\ \hline
Ours                          & 2016-2020       & 157                 & Any                & Any              \\ \hline
\end{tabular}

\caption{Related Reviews}
\label{tab:relatedreviews}

\end{table}

\subsection{Classification Rationale for Computer Vision Approaches}

In this paper we roughly divide the analyzed papers, from the point of view of the image processing technology used, into three general categories: (a) classic CV approaches (CCV), (b) deep learning (DL) approaches, and (c) hybrid CV approaches (HCV), where DL and CCV are employed together. 

The general principle behind the CCV paradigm is that an image interpretation task starts from meaningless pixels and moves stepwise towards a
meaningful representation of that image's content. The transform steps move towards a growing abstraction and simplification of the image to find and isolate the elements of the image that have a meaning for a given particular application context. This is performed as a pipeline of transforms in different \emph{levels of abstraction}, whereas the first transforms in this pipeline are transforms from images into new and simplified images $(I \to I )$, such as noise filters, border detectors or segmentation algorithms. Later, when the abstraction level grows, these transforms are performed from images into models $(I \to M )$ that represent descriptions of specific elements or the content of these images, such as segment color and shape parameters or texture descriptors. Further on, these models are transformed into other, more abstract models $(M \to M)$ that describe the meaning or classification of those objects \cite{Marr1982}. These transform steps can be applied on three different \emph{image processing domains}: (i) \emph{value}, when only context-less individual pixel values are being considered, such as in histograms or thresholds; (ii) \emph{space}, when pixels are considered spatially, in the context of their neighborhoods, such as in border detectors, segmentation methods or general convolutions; and (iii) \emph{frequency}, when pixel variation patterns along the image are taken into consideration, such as in Fourier or Wavelet transforms \cite{Marr1982}.  

One characteristic of CCV is that these pipelines are extremely problem- and image type-specific and strongly parameter-dependent. This means that a processing pipeline that works well for a given type of image content, e.g., Papanicolaou samples, will not work for Feulgen-stained samples. This parameter sensitivity also means that, e.g., a segmentation algorithm that works well for a set of images, will not work as well with another set of images of the same kind, but that shows different lightness and contrast characteristics. This made CCV solutions challenging to develop, requiring deep mathematical knowledge of the individual methods in order to allow an adequate composition of a pipeline for a specific problem. These characteristics also made CCV solutions less robust and highly image-quality-dependent, which has led to some image processing cytology solutions to require the user to perform a rigorous process for the preparing and digitizing of the samples, to guarantee that the image processing pipeline actually works \cite{Boecking2019}.  

\begin{sidewaysfigure}
    \centering
        \includegraphics[width=1.1\textwidth]{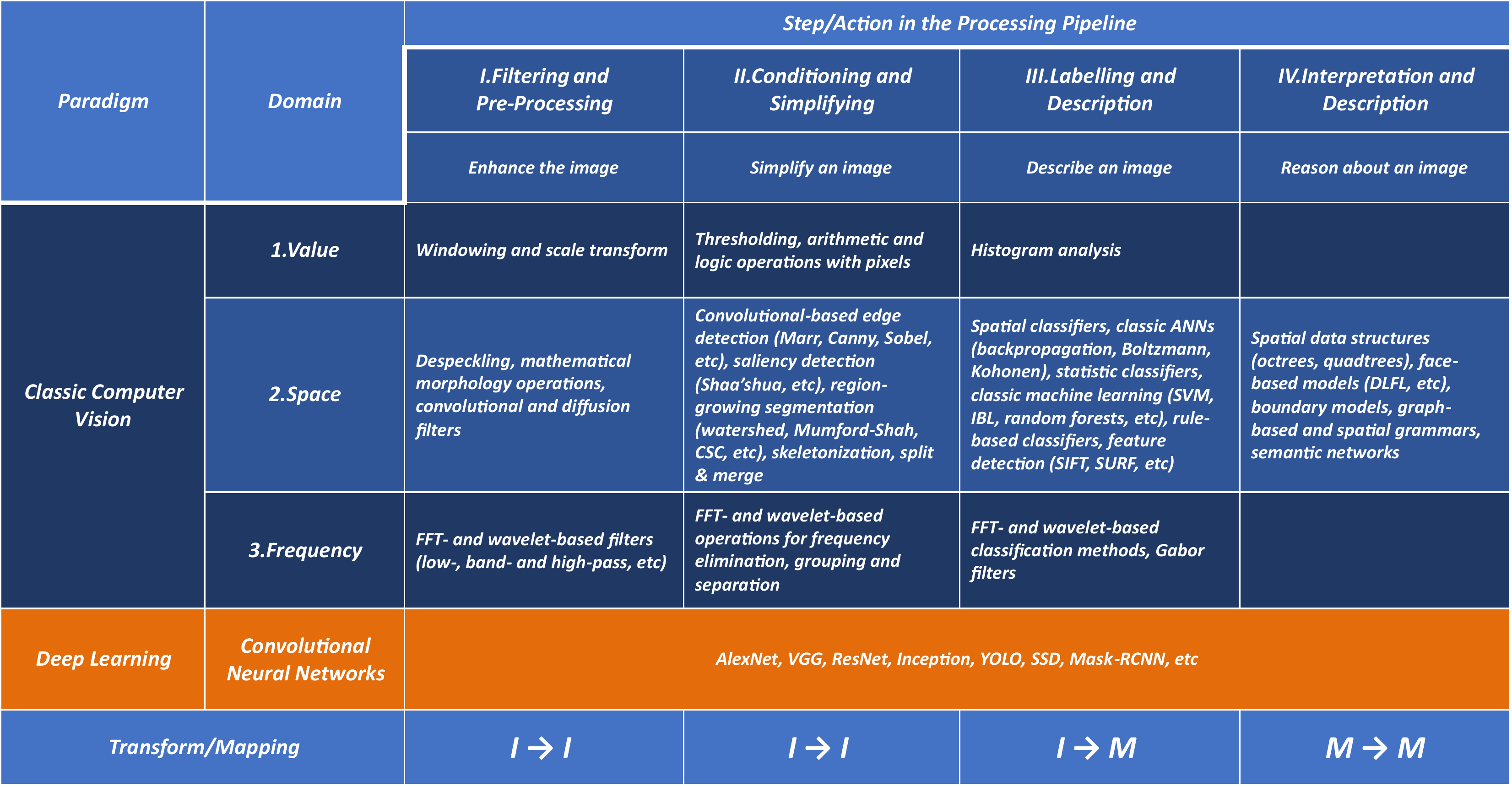}
        \caption{Computer vision domains, steps and paradigms.}
        \label{fig:domains}

\end{sidewaysfigure}

The advent of DL techniques broke the CCV paradigm: DL convolutional neural networks (CNN) can learn sequences of convolution operations that represent image transforms in all three image domains and all levels of abstraction \cite{LeCun2015}. Due to their deep sequential structure, DL CNNs can also learn hierarchical transformation and representation sequences \cite{LeCun2015}. A DL application is able to, in one single step, start from pixels and end with a representation of the meaning of a given image. DL also simplified the development process of CV applications: the main steps of a CV solution can now be \emph{trained}. However, DL solutions have the disadvantage of being black-box processes: the convolution operations performed by the network are learned and coded as data in the structure of a given network, and not separated processes that can be isolated and individually analyzed. 

An overview of both paradigms is shown in Figure \ref{fig:domains}. For this work, we considered CCV approaches all those that did not use any DL technique in its processing pipeline.

 \subsection{Staining Techniques} \label{staining-techniques}

\begin{figure}[H]
        \begin{center}
            \includegraphics[width=0.8\linewidth]{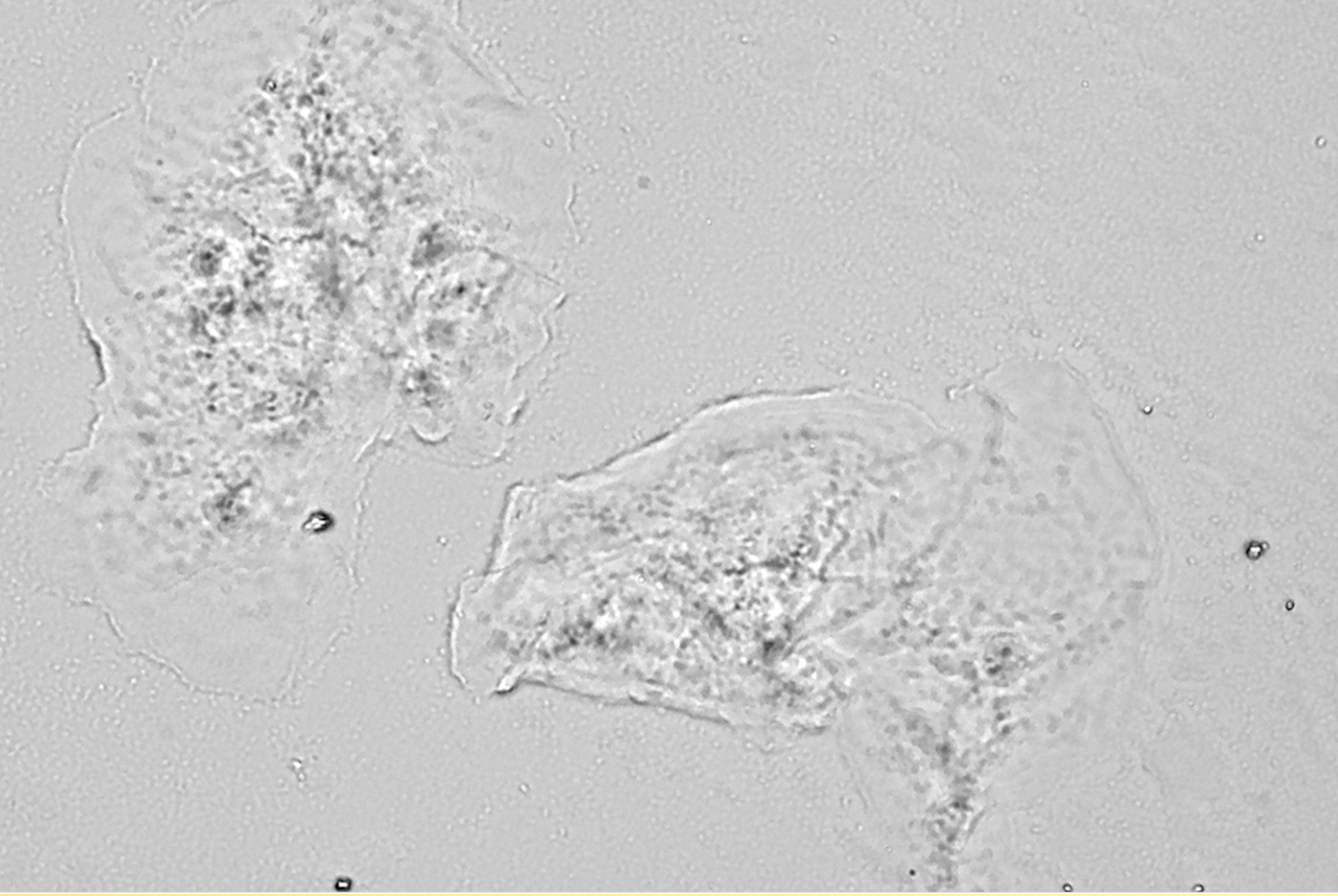}
            \caption{Example of an unstained cytology slide}
            \label{fig:unstained-example}
        \end{center}
    \end{figure}

\begin{figure}[H]
        \begin{center}
            \includegraphics[width=\linewidth]{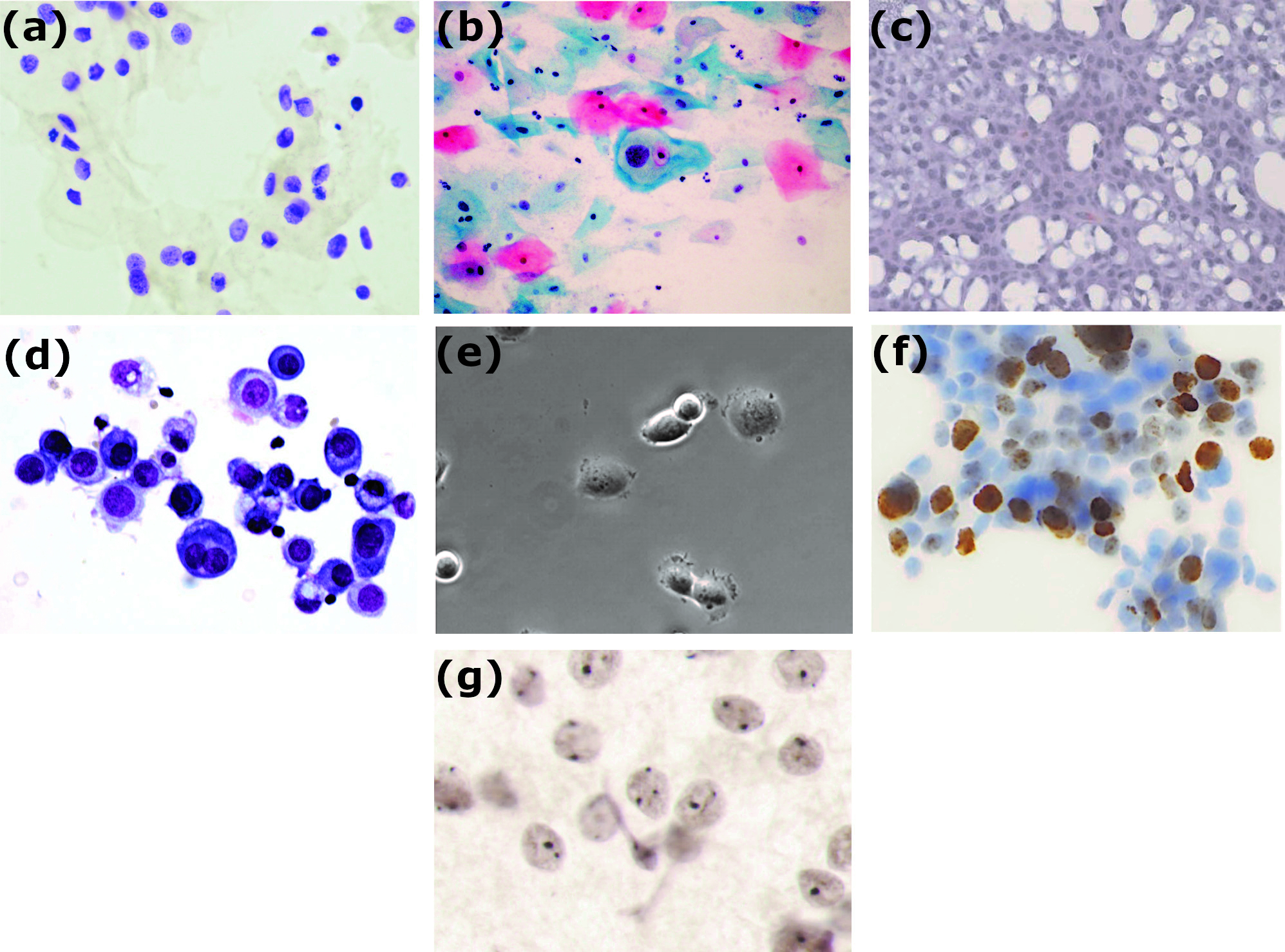}
            \caption{Examples of staining techniques. (a) Feulgen, (b) Papanicolaou, (c) H\&E \cite{Tsukada2017}, (d) Giemsa, (e) Unstained using phase-contrast microscopy \cite{Hu2017}, (f) Hematoxylin+DAB \cite{Kowal22018}, (g) AgNOR}
            \label{fig:stainings-example}
        \end{center}
    \end{figure}

    Cytology examinations are performed through the use of microscopes to evaluate slides prepared from small sets of cells. Usually, a staining technique must be used to highlight the cells' desired features.
    Figure \ref{fig:unstained-example} shows part of an unstained slide and Figure \ref{fig:stainings-example} shows an example of the results of applying different staining and highlighting techniques that we have identified in the analyzed papers. The techniques applied in these papers are Papanicolaou (130 papers), H\&E (20 papers), Giemsa (5 papers), Feulgen (5 papers), Hematoxylin+DAB (1 paper), AgNOR (1 paper), and 1 paper that employed unstained slides. There are also 3 papers that do not specify clearly which staining method was used.

    Except for \cite{Hu2017}, research described in all papers employed some kind of staining technique, with Papanicolaou being the most frequently used. Papanicolaou is an inexpensive staining technique, generally applied to detect cervical cancer, composed of five dyes (Hematoxylin, Orange G, Bismark brown Y, light green SF yellowish, and eosin Y) that aims to define nuclear details, highlight cell boundaries, and differentiate between types of cells. Although it is widely used, the Papanicolaou staining is not a well-standardized method, which produces a considerable range of variations in its results \cite{Gill2013}. Hematoxylin is a dye that stains the chromatin and is generally used along with eosin (H\&E), a dye that stains the nucleoli and the cytoplasm of the cells \cite{Gill2013}. \cite{Krau2017} states that Raman spectroscopy is more effective than the use of H\&E staining for feature extraction. One of the analyzed papers \cite{Kowal22018} used DAB (3,3'-Diaminobenzidine) instead of eosin for immunohistochemical staining. The Giemsa stain is one of the Romanowsky-type stains, which allows the estimation of relative cell and nuclear sizes and the visualization of cytoplasmic details, smear background elements, and intercellular matrix components \cite{Gill2013}. Feulgen staining is used to identify chromosomal material or DNA in the cells specimens \cite{Gill2013} and the AgNOR technique consists of applying a silver nitrate staining to the slide to highlight \ac{NORs} \cite{Derenzini2000}.

\subsection{Datasets} \label{datasets}
Data plays an essential role in machine learning, particularly for deep learning approaches \cite{macarini2020complete}. When dealing with medical data segmentation, the data availability is often a problem \cite{eaton2018improving}, sometimes being necessary to collect and label more samples. Publicly available datasets are important tools for benchmarking, providing means for a fair comparison between the methods published in the literature.

We were able to identify 12 publicly available datasets: HEMLBC \cite{Zhang2014}, ISBI 2014 \cite{ZhiLu2015}, ISBI 2015 \cite{Lu2017}, DCCL \cite{Zhang22019}, Herlev \cite{86c7f5c1a9f84a7484731dd71671c563}, SIPaKMeD \cite{plissiti2018sipakmed}, Perolehan Citra Kanker Serviks \cite{Serviks:2020}, AGMC-TU Pap-Smear \cite{Bhowmik2018}, CCAgT: Images of Cervical Cells with AgNOR Stain Technique \cite{AtkinsonAmorim2020, CCAgT2020}, UFSC-OCPap \cite{Matias2020},
and 2 unnamed datasets proposed in \cite{Saikia2019} and \cite{Arajo2019}. Except for HEMLBC, which consists of Liquid-based H\&E slides, and the CCAgT, which consists of AgNOR slides, all the other datasets use the Papanicolaou staining technique. The DTU/Herlev dataset was used in 38 works, being the most recurrent among all datasets. 

Ten of the twelve public datasets found are composed of samples collected from the Cervix, except for the dataset presented in \cite{Saikia2019}, which is composed of \ac{FNAC} Smears collected from the breast, and the UFSC-OCPap dataset, which is composed of oral smear samples.

\subsection{Image Classification} \label{image-classification}

Image classification is a fundamental task in computer vision. This task aims at categorizing images into one or several predefined classes. Image classification can be considered the basis for other computer vision tasks, e.g., localization, detection, and segmentation \cite{rawat2017deep}. Even though it is a trivial task for human beings, it is challenging for an automated system \cite{ciresan2011flexible}. In the past few years, the developers usually adopted a two-step image processing pipeline approach to solve this problem. First, a set of image type-specific features were extracted from images using a sequence of one or more handcrafted feature descriptors. Afterwards, these features were used to train a classifier \cite{lecun1998gradient}. However, there were many complications when using this approach, such as viewpoint-dependent object variability and the high intra-class variability of having many object types \cite{ciresan2011flexible}. One of the main problems of this approach was the fact that the accuracy of the classification task was profoundly dependent on the design and robustness of the specific feature extraction algorithms employed \cite{lecun1998gradient}. Nowadays, deep learning models are being used for integrated feature extraction and transformation, as well as for pattern analysis and classification, integrating all steps of this image processing pipeline into a single large classification step performed by a deep learning neural network. It has been shown to overcome these challenges \cite{rawat2017deep}. 

\subsubsection{Classic Approaches}

In most of the papers we retrieved, classic CV approaches are used to extract features of the images for later classification using the \ac{ANN}, \ac{SVM} or \ac{kNN} techniques. In \cite{Rasche2017}, however, the authors employ an approach based only on the contours of the nucleus, where the classifier uses the geometric features and the \ac{FFT} to describe this external contour. 
 
For the feature extraction step, multiple approaches were employed: Histogram analysis, Gray level Co-occurrence matrix (GLCM), Local binary pattern (LBP), Laws texture energy-based, discrete wavelet transform (DWT), Markov Random Fields, morphometric, textural and intensity. In \cite{lin2017llc}, the authors used Scale-invariant Feature Transform (SIFT) for the feature extraction step. Classification using \ac{SVM} was used in \cite{Banerjee2016, GarciaGonzalez2016, Zhao2017, Arya2018, Bora2017, Wang2019, Zhao2016, AnousouyaDevi2018, lin2017llc, chankong_automatic_2018, Somasundaram2020}, \ac{ANN} in \cite{Kar2019, GarciaGonzalez2016, Arya2018, Makris2017, Sunny2019, Bora2017, han2018suspected, dumripatanachod2016centralized, fekri-ershad_pap_2019}, Decision Tree with Logistic Regression in \cite{Su2016} and \ac{RF} in \cite{Boecking2019, Bora2017, chankong_automatic_2018}. In \cite{RodrguezVzquez2017}, the authors used \ac{kNN}, \ac{SVM}, and \ac{ANN} to extract features from manually annotated cytoplasm and nuclei. In \cite{Bhowmik2018}, the authors published their dataset (AGMC-TU Pap-Smear), where they perform a benchmark among models of \ac{ANN}, \ac{SVM}, \ac{kNN}, \ac{LDA}, \ac{RF}, decision tree and Naive Bayes to classify the pap-smears cells in normal and abnormal. Among these papers, 2 work with oral cavity samples, 7 use cervical Pap smear samples, and in \cite{Boecking2019} the authors use lung cells slides stained with Feulgen. In \cite{deMesquitaSJunior2018}, the authors propose using the Randomized Neural Network Texture Signature method to classify cells in Pap smear slides. Global Significant Value, statistical-based texture features and interquartile range were used in \cite{fekri-ershad_pap_2019} to extract cell features used as input to a Naive Bayes model, a J48 tree and an \ac{ANN} for classification. In \cite{abhinaav_abnormality_2019}, cell features were classified into normal and abnormal using two classifiers: a two-class boosted decision tree and a two-class logistic regression model. The best result between the classifiers was passed to multi-class logistic regression classifiers to classify the abnormal cells into 4 classes. In \cite{ramdhani_hierarchical_2017}, the authors compare an \ac{ANN}, combined with a genetic algorithm for feature optimization, and a Hierarchical Decision Approach (HDA). The results indicate that the HDA had the best results.
The work from \cite{chankong_automatic_2018} uses the 2-D Fourier transform to extract spectral features that are used to train the following classical machine learning algorithms: \ac{kNN}, \ac{SVM}, Random Forest, and AdaBoost.
 
\subsubsection{Deep Learning Approaches}

The \ac{DL} approaches we retrieved employ the most recent deep convolutional neural network models seen in the state-of-the-art CV research, such as Residual Neural Network (ResNet) models \cite{he2015deep} and VGG net \cite{simonyan2014deep}.

In this sub-category, we were able to identify research works employing five different stains and cells from four different anatomies. In \cite{Hu2017}, the authors used a \ac{CNN} model to select single cell images in the first stage, and in the second stage, a \ac{CNN} based on \textit{VGG\_CNN\_F} \cite{chatfield2014return} to classify human bladder cancer cells was used. In \cite{nguyen2018deep}, the authors proposed a novel deep neural network to classify microscopy images. They concatenate the features extracted using three pre-trained deep CNNs. These features are used to train two fully-connected layers to perform classification.

In \cite{Xiang2020}, the authors have used cervical slides stained with Feulgen to realize a manual feature extraction and classify the cells into 10 categories using the InceptionV3 \cite{szegedy2016rethinking} model. The only \ac{DL} classification work that used images of nasal mucosa cells with the May Grunwald-Giemsa (MGG) stain was \cite{Dimauro2018}. They have used a small \ac{CNN} with two layers proposed to classify the cells into seven distinct classes. We also identified two papers that used the H\&E-stain. In \cite{Bianconi2019}, the authors trained four models: VGG-16, VGG-19, ResNet-50 and ResNet-101. The ResNet-101 was the model described as being the one that showed the best overall performance.

In \cite{Garud22017}, the authors also employed H\&E staining with breast \ac{FNAC} samples, which were classified using the GoogLeNet \cite{Szegedy2015} model. In this paper, the authors propose a voting pre-processing to select the fields of the slide used as input in order to classify it. In \cite{elliott2020application}, the authors developed an algorithm to evaluate thyroid fine-needle aspiration biopsy using whole slide images. The designed algorithm is based upon two CNNs: one to identify the follicular groups and another to predict the Bethesda System for Reporting Thyroid Cytopathology category and the final pathology. Both networks are based on the VGG-11 model, pre-trained on the ImageNet dataset. According to the authors, the performance of the algorithm was comparable to that of an expert cytopathologist. In \cite{gupta2020region}, the authors proposed a system that displays regions of interest on whole slide images containing potential anomaly. Then, the pathologists review these regions to increase productivity. The authors built multiple deep learning classifiers, using VGG-16 and DenseNet-121, as part of the proposed architecture. 

We analyzed more than ten papers that employed the Papanicolaou staining technique. To classify lung cells, the authors used a model based on VGG-16 to classify pulmonary images in \cite{Teramoto2019}. \cite{Baykal2019} compared the ability of AlexNet, GoogLeNet, ResNet, Inception V3, VGG and DenseNet to classify the cells into eleven classes. In \cite{lin2019fine}, the authors compared AlexNet, GoogLeNet, ResNet, and DenseNet to classify nuclei in the Herlev dataset. A Deep Belief Network made of Restricted Boltzmann Machines was proposed in \cite{Rasche2017}, where the model was compared to a classic approach. There are 19 papers that showed similar approaches with different models to classify Pap Smear images, where \cite{gv_automatic_2019, Zhang22019, Deng2020, Khamparia2020, Mousser2019, pmid32308878, lu2019deep, Matias2020} used the ResNet model, \cite{Zhang22019, Khamparia2020, Mousser2019, pmid32308878, Gonzalez2020} used the Inception model, \cite{Khamparia2020, Mousser2019, pmid32308878} used the VGG model, \cite{Zhang22019, pmid32308878} used the DenseNet model, \cite{Hussain2020} used a combination of the ResNet model and the DenseNet model, \cite{Rasche2017} used a Deep Belief Network, \cite{Martinez2020, Taha2017} used the AlexNet model, \cite{Khamparia2020} used the SqueezeNet model, and \cite{Ke2020, Shanthi2019, NirmalJith2018, Aljakouch2019, Zhang22017, Sanyal2019} used custom CNN models.

The approach used in \cite{Deng2020} was divided into two steps: In the first, Pap Smear images were segmented into nuclei, cytoplasm and background, employing a U-net model using a ResNet as its backbone. In the second step, the nuclei were classified into sub-types. Combining a \ac{CNN} with multiple instance learning, the authors in \cite{Ke2020} proposed a model for dysplasia detection. The authors of \cite{Khamparia2020} and \cite{Bora2016} propose a two-step method where a \ac{CNN} is used to extract the image features that are then passed to a classifier to produce the final result. In \cite{Khamparia2020}, they compare VGG-19, SqueezeNet, Resnet-50, and InceptionV3 to extract the images features and \ac{kNN}, \ac{RF} and \ac{SVM} for classification. In \cite{Zhang22019}, the models ResNet-101, InceptionV3, and DenseNet-121 were compared to classify cells in Pap Smear images. In \cite{Hussain2020}, for pre-cancer lesions classification, the models AlexNet, VGG-16, VGG-19, ResNet-50, ResNet-101, and GoogLeNet were compared. The work from \cite{wang_adaptive_2020} uses a network composed of 10 convolutional layers and three fully connected layers pre-trained on ImageNet. Then, the proposed method discards all the non-relevant kernels according to a score and the network is trained using a private cervical pap smear dataset.

\subsubsection{Hybrid Approaches}

We considered as Hybrid approaches the ones that employed classical image processing/computer vision methods for pre-, post-processing or image enhancing, for example, together with a DL step for image classification. In \cite{Saikia2019}, the authors considered the cytoplasm and red blood cells present in the images as noise. Then, Histogram equalization was employed in the red channel output as a pre-processing step to remove the cytoplasm. Afterwards, images were segmented to remove the cytoplasm and the red blood cells. The global and the Otsu thresholding were employed to make it simple. According to the authors' visual investigation, Otsu thresholding performed better than global thresholding. This method was used to isolate the nuclei in each image, and then the authors compared four different \ac{CNN} models for the classification of \ac{FNAC} samples.

In \cite{Dimauro2019}, the authors proposed a three-layer \ac{CNN} to classify nasal cytology   samples. In the pre-processing step, aiming to create a system capable of recognizing images of slides prepared with different techniques, the Otsu algorithm was applied in grayscale images. The morphological dilation and the Watershed algorithm were applied. Then, the next step is labelling, for marking the different objects with different shades of colour to facilitate subsequent classification. In \cite{plissiti2018sipakmed}, the authors presented a new database of cervical images where the cells were categorized into 5 classes. Among other classic methods, the authors employed a VGG-19 for cell classification. The work from \cite{faturrahman_multi_2017} extracts features from the shape of the cell and use additional methods as Local Binary Pattern (LBP), Gray Level Co-Occurrence Matrix (GLCM) and Shape Features to extract additional features which were later used to train multiple Deep Belief Networks that were developed using multiple combinations of hidden layers and hidden units to perform the experiments and find out which is the best configuration. 

\subsection{Object Detection}

Object detection is a computer vision task that aims to localize each instance of certain object classes in an image. It can be seen as a combination of two concepts: the first, image classification, which consists of predicting the class of one object in an image; the second, object localization, refers to identifying the location of one or more objects in an image. Object detection combines these two tasks to localize and classify one or more objects in an image. Since the advent of \ac{CNN}s, object detection has experienced impressive progress in terms of accuracy and speed.

Concerning object detection applied to cytology, we could identify that all the papers employed directly state-of-the-art object detectors. None of the authors proposed a new architecture of \ac{CNN}. Among the different architectures used, two works used the YOLOv3 model. In \cite{Xiang2020}, the authors used YOLOv3 \cite{yolov3} to detect cells in images collected from the cervix. A task-specific cascade classifier was used to improve the classification performance of hard examples. In \cite{Kilic2019}, the YOLOv3 was used to detect cell nuclei on pleural effusion cytopathology images stained with Papanicolaou. According to the authors, the most significant contribution of the method was the detection speed.

In \cite{Zhang22019}, it was presented, according to the authors, the largest cervical cytology dataset, called Deep Cervical Cytological Lesions (DCCL). The authors used a Faster R-CNN and a RetinaNet \cite{retinanet} with a ResNet-50 backbone as baseline detection models. For classification, the authors used three different CNN architectures: InceptionV3, ResNet-101, and DenseNet-121. In \cite{Du2019}, a region detection and classification method based on multi-semantic label combined with morphological information analysis was proposed. The authors then used transfer learning to train a Faster R-CNN model, with VGG-16, ResNet50, and ResNet101 as backbones to classify the dataset. 

The authors from \cite{Baykal22019} also used the Faster R-CNN, R-FCN, and SSD architectures for nuclei detection in pleural effusion pathology images. They used different feature extractors, namely, ResNet, Inception v2, and MobileNet for performance comparison. In \cite{Khan2019}, the authors proposed a framework to extract features from H\&E stained images using pre-trained CNN architectures: GoogLeNet, VGG, and ResNet. The feature extractor output is then fed to a fully connected layer to classify malignant and benign cells using average pooling classification. The authors from \cite{Su2020} tested multiple detection network architectures applied to H\&E and Papanicolaou stained slides to detect peritoneal metastasis in ascites cytopathology. In \cite{Sompawong2019}, the authors used a Mask R-CNN model for nuclei detection in Pap smear slides and, in \cite{lu2019deep}, an \ac{FCRN} is used to detect nuclei in cytology slide images that are after used to test ResNet50 and DenseNet201 for single-cell nuclei classification. The authors from \cite{Matias2020} compared the Faster R-CNN and the RetinaNet models, using ResNet-50 as the backbone, to detect cell nuclei in Papanicolaou stained slides using both a binary and a multiclass approach. 

Besides most of the detection papers analyzed use a \ac{CNN} as the base model, in \citet{Mufidah2017} the authors employed an \ac{SSAE} along with a softmax classifier to detect nuclei in patches extracted using sliding windows. This method was tested to detect cell nuclei in Pap smear images containing overlapped cells.

\subsection{Segmentation-Based Approaches}

Segmentation techniques in cytology aim to provide each pixel of an image with a label, classifying them between desired (e.g. cells or nuclei) and undesired objects (e.g. background), or between multiple classes (e.g. background, normal cell, abnormal cell, etc.), in order to interpret the image. 

\subsubsection{Classic Approaches}

Classic segmentation approaches are based on operations over pixel intensity values, generally in the context of their neighborhoods, where the input usually is a matrix of color values that is processed to generate a matrix of the pixels' labels as the output. In the analyzed papers, this approach is used for detection or counting cells in 23 papers, segmentation of overlapping cells in 19 papers, and features extraction for classification in 40 papers. Due to the high dimensionality of the extracted features, 15 papers use a dimension reduction algorithm before passing the data to a classifier. Most of the selected features are based on the morphology and texture of the cells, although some papers use cell distribution information and \cite{Jele2016} states that “moment-based features offer better separation between the malignancy classes than shape and texture based features”. The tools used for classification of the extracted features are \ac{SVM} (20 papers), \ac{RVM} \cite{Taneja2017}, \ac{ANN} (11 papers), \ac{kNN} (6 papers), Naive Bayes (3 papers), Random Subspace-Linear Discriminant Analysis \cite{Nawandhar2020}, k-means (2 papers), Fuzzy C-Means (5 papers), \ac{HOG} \cite{prum_abnormal_2018}, Decision Tree (8 papers), Random Forest (9 papers), Logistic Regression (2 papers), \ac{LDA} (3 papers), and \ac{NCA} \cite{Nawandhar2020}. Also, in \cite{Win2019} and \cite{mishra2018theoretical}, the authors use an ensemble classifier that combines the results of multiple algorithms to generate the final output using major voting.

To apply the segmentation processes, it is usually necessary to pre-process the image to highlight the desired features and smooth or eliminate undesired ones. The pre-processing techniques used in the selected papers were: conversion of the images’ RGB color space to the grayscale (24 papers),  HSV (5 papers), and l*a*b* (7 papers) color spaces, besides noise reduction (32 papers), contrast adjustments (13 papers), convolutional filters (7 papers), gamma correction \cite{SAHA22017}, and histogram equalization (7 papers). The pre-processed image is then passed to the actual segmentation algorithms, which are divided into three groups in this section: thresholding-based, trainable model-based, clustering-based and graph-based. Papers that use thresholding-based techniques (44 papers) apply a local or global threshold over the input image to divide its pixels between the background and the desired objects and to detect contours. Clustering-based techniques group the pixels based on a similarity measure using clustering algorithms like k-means (12 papers), \ac{DBSCAN} (2 papers), Fuzzy C-means (4 papers), \ac{SLIC} superpixel merging (6 papers), Mean shift \cite{Wang2019}, and Gaussian mixture model (3 papers). Graph-based approaches treat the image pixels and their relations as a graph and employ graph algorithms to segment the image, as shown in 6 papers. Trainable model-based methods use pixel-level information to train a model to distinguish between the regions of interest in the image and the background. These methods were used in 4 of the analyzed papers.

Moreover, these techniques usually rely on additional mathematical models, such as mathematical morphological operations (24 papers), active contours (7 papers), geometric shape constraints  (17 papers), and other techniques (4 papers). These mathematical models are employed in the post and/or pre-processing phase to optimize the segmentation results. A table with details of each paper that apply classic segmentation-based methods is presented in \ref{tab:papers-segmentation-classic}.

\subsubsection{Deep Learning Approaches}

In this section, we will describe papers that employed Semantic Segmentation and other CNN-based segmentation approaches. Semantic Segmentation (SS) is a CNN technique where each pixel of an input image is individually classified and, thus, the image is segmented into different parts and objects, grouping pixels that \emph{semantically} belong together, i.e., belong to the same class of the GT used to train the CNN \cite{Guo2017}.  
The most widely used network architectures to solve segmentation problems are based on the U-Net \cite{Ronneberger_2015} and SegNet \cite{Badrinarayanan_2017} architectures. Both models use only convolutional layers and consist of \emph{encoders} and \emph{decoders}. The image that enters the network is transformed into a feature map and each encoder progressively reduces the size of these maps by condensing the features. The decoders do the opposite process: they take the output from the last encoder and progressively expand the feature maps, turning them into image segments.

Most \ac{DL} papers \cite{Zhang2017, Tsukada2017, Ke2019, Das2019, Arajo2019, Deng2020, BaykalKablan2020, allehaibi2019segmentation, song2016segmenting, sabeena_improved_2020} applied classic neural networks like VGG, AlexNet, U-Net, Mask R-CNN, and ResNet to segment cells. Some of these papers applied extra steps to improve the segmentation accuracy. An SVM classifier was used in \cite{Tsukada2017} to predict the nuclei presence probability in slide images. A voting ensemble method combining a fully convolutional network, a SegNet, and a U-Net applied to a \ac{CNN} was used in \cite{BaykalKablan2020}. A double U-Net, as known as W-Net, was used in \cite{Das2019}.

Some papers used non-conventional neural networks with additional steps. The DeepLabv2 network with a modified U-Net, called TernausNet, was used in \cite{Wan2019} and in \cite{Zhao22019} a modification of the U-Net is proposed, called PGU-net+, where a residual module is added to the model and it is trained using progressive growing input sizes. Cascade regression chain was used with a custom neural network in \cite{Song2018}. A binary tree-like network with two-path fusion attention failure was used in \cite{Zhang2019}. The authors from \cite{Hussain20202} implemented a Fully Convolutional Neural Network (FCN) model based on the U-Net. In \cite{zhou2019irnet}, the authors presented the IRNet, a novel instance relation network for overlapping cell segmentation. The authors from \cite{AlemiKoohbanani2020} propose a model based on the U-Net with residual blocks to use as a framework to assist in the segmentation of nuclei, cells, and glands. They trained the model in histopathology datasets and validated it using a Pap smear dataset, showing that it can also segment cells in cytology images. In \cite{lauw_mird-net_2020}, a network based on the U-Net without the pooling layers combines the features from the inception network with dense blocks to make the network obtain more features with fewer parameters. The work from \cite{zhao_automated_2019} applies a network based on the U-Net that uses a deformable convolution to improve the detection of the irregular shape of nuclei in abnormal cells. The ensemble is made using 3 different neural networks: Dense U-Net, Dense U-Net with deformable convolutions on the expansion path and a Dense U-Net with deformable convolutions on the contracting path. The authors from \cite{yang_interacting_2020} apply an encoder and decoder architecture, like the U-Net, with a two-path parallel convolutional module. The module consists of a small kernel to extract features from inside the nucleus and a large kernel to extract features from the whole cell. Also, a Feature Pyramid Network is applied to reduce the information loss on feature maps caused by the pooling layers. In \cite{AtkinsonAmorim2020}, the authors use the U-Net model for segmentation of Argyrophilic Nucleolar Organizer Regions (AgNOR) by Silver staining technique, classifying the nuclei and the NORs as clusters and satellites. A similar method is used in \cite{Matias2020} for nuclei classification in Papanicolaou images, where the U-Net model is applied using ResNet-34, ResNet-50, and ResNet-101 as backbones in multiple image resolutions using a binary and a multiclass approach.

\subsubsection{Hybrid Approaches}

We classified as hybrid approaches the works that use neural networks in combination with classic CV methods to segment regions of interest. The Watershed method was used along with a neural network in \cite{Win22017} and \cite{Kowal2018}. This method consists of pixels chosen as seeds, which then aggregate with the other pixels into \emph{basins} according to pre-set criteria such as gray level, color, and texture. 

A superpixel-based method with neural networks was applied in 3 papers. A Voronoi diagram generated with a Delaunay triangulation was used in \cite{Tareef2017}. The Delaunay triangulation divides the plane into multiple triangles, with the points as vertices, and the Voronoi algorithm uses those points to create areas where boundary lines are equidistant from neighboring points. A Multiscale Laplacian of Gaussian was used in \cite{Xu2018}, where the algorithm detects the edges of an image varying the size of the image. The paper \cite{Zhang32017} tested multiple generic segmentation methods like threshold, Hough transform, and deep learning methods as a first segmentation step, using a graph-based method to optimize the results. The approach from \cite{harangi_cell_2019} combines a superpixel segmentation method with two Fully Convolutional Neural Networks to create an ensemble that merges the outputs from those methods. The networks used in the experiment were based on VGG-16.

\section{Results}

    In this SLR, our goal was to discover the state-of-the-art of CV methods employed in computer-aided diagnosis using cytological images. We searched the Springer, IEEE, ScienceDirect, ACM, Wiley, and PubMed databases. Our research returned 431 works from which, after applying the inclusion/exclusion criteria, remained 157 papers that we read and analyzed. During the analysis of the works, some main points of interest were mapped: the processing image method utilized, which part/organ of the body and pathology was analyzed, the stain utilized, the availability and size (considering the number of patients and images) of the used dataset, the training strategies, and the evaluation metrics applied.
    
    Regarding the classic approaches, most of these methods are in the Space domain. The most known methods are GLCM, DWT, LBP, Watershed, and Hough Transform. There are less usual methods like  Laws Texture Energy-based, Markov Random Fields, Voronoi Diagram with Delaunay Triangulation, and Multiscale Laplacian of Gaussian. For segmentation, we identified some clustering-based techniques like k-Means Clustering, \ac{DBSCAN}, Fuzzy C-Means, \ac{SLIC}, Superpixel Merging, Gaussian Mixture Model, and the Active Contours. There are also the morphological operations that are used, usually, in pre- and post-processing steps.  In the Value domain, we could identify methods like Histogram Equalization, OTSU, and global thresholding. The less recurrent methods are the ones present in the Frequency domain, like DWT and FFT. For classification, we identified the ANN, kNN, Random Forest, Relevance Vector Machine, Naive Bayes, Decision Trees, Neighborhood Component Analysis (NCA), Random  Subspace-Linear Discriminant  Analysis, logistic regression model, Hierarchical Decision Model (HDA), AdaBoost, Relevance Vector Machine (RVM),  Random Subspace Linear Discriminant Analysis, Neighborhood Component Analysis, and the most recurrent, the SVM.

  For deep learning, most of the networks identified in this work are an application of unmodified architectures well established in the literature. For example, we could identify: VGG-11, VGG-16, VGG-19, Inception V2, Inception V3, ResNet50, ResNet101, GoogLeNet, AlexNet, DenseNet, SqueezeNet, RetinaNet and Deep Belief Networks. For object detection, we could identify YOLOv3, Faster R-CNN, MobileNet, and SSD based networks. We identified two novel architectures, the W-Net (double U-Net), and the TernausNet, composed of a modified U-Net and the DeepLabV2. For semantic segmentation, we identified the SegNet, and the U-Net, being the most used.

\begin{figure}[htp!]
        \begin{subfigure}[b]{.5\textwidth}
          \centering
          \includegraphics[width=\linewidth]{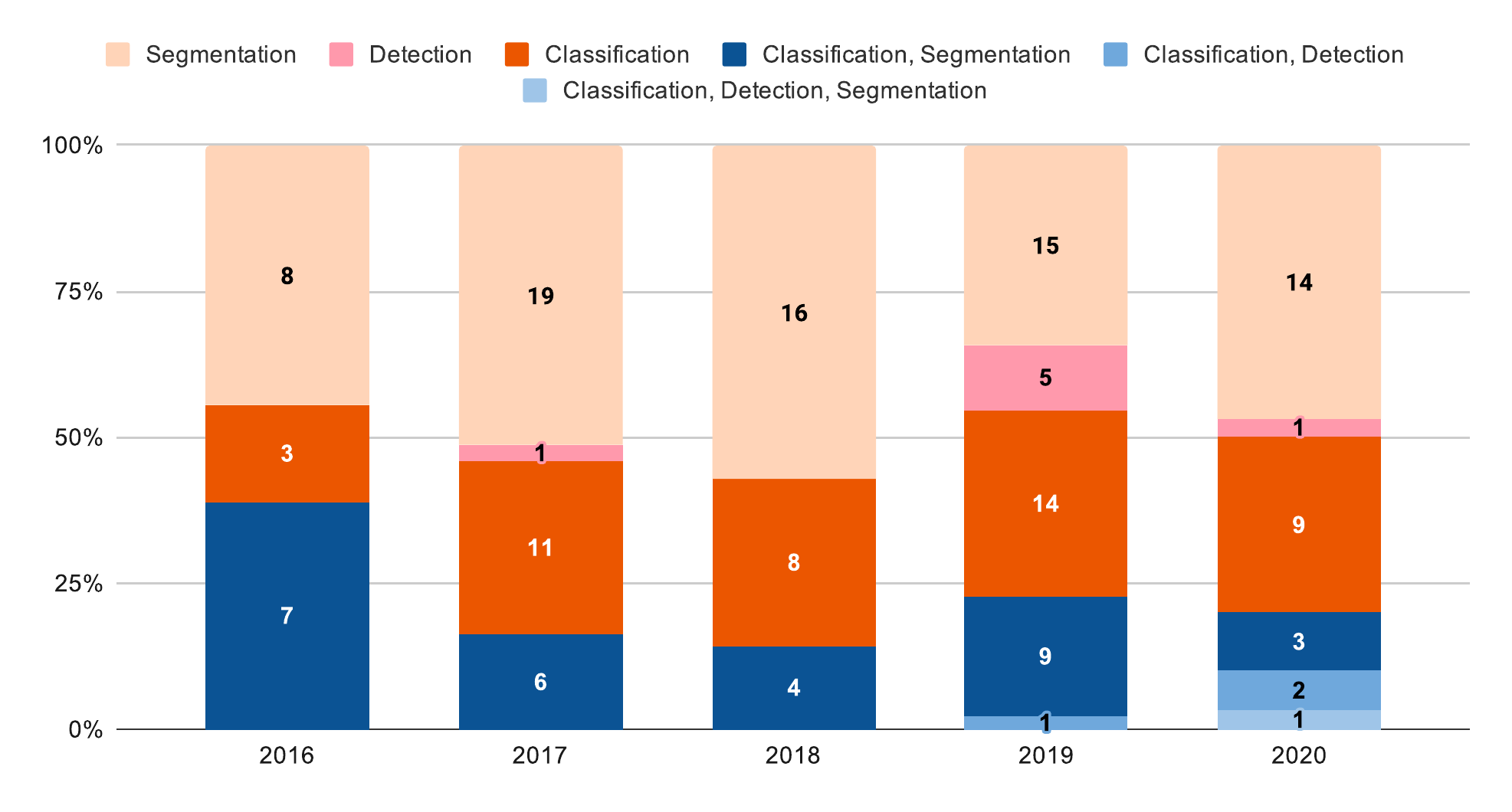}  
          \caption{Methods by year}
          \label{graph:papersbyyear}
        \end{subfigure}\quad
        \begin{subfigure}[b]{.5\textwidth}
          \centering
          \includegraphics[width=\linewidth]{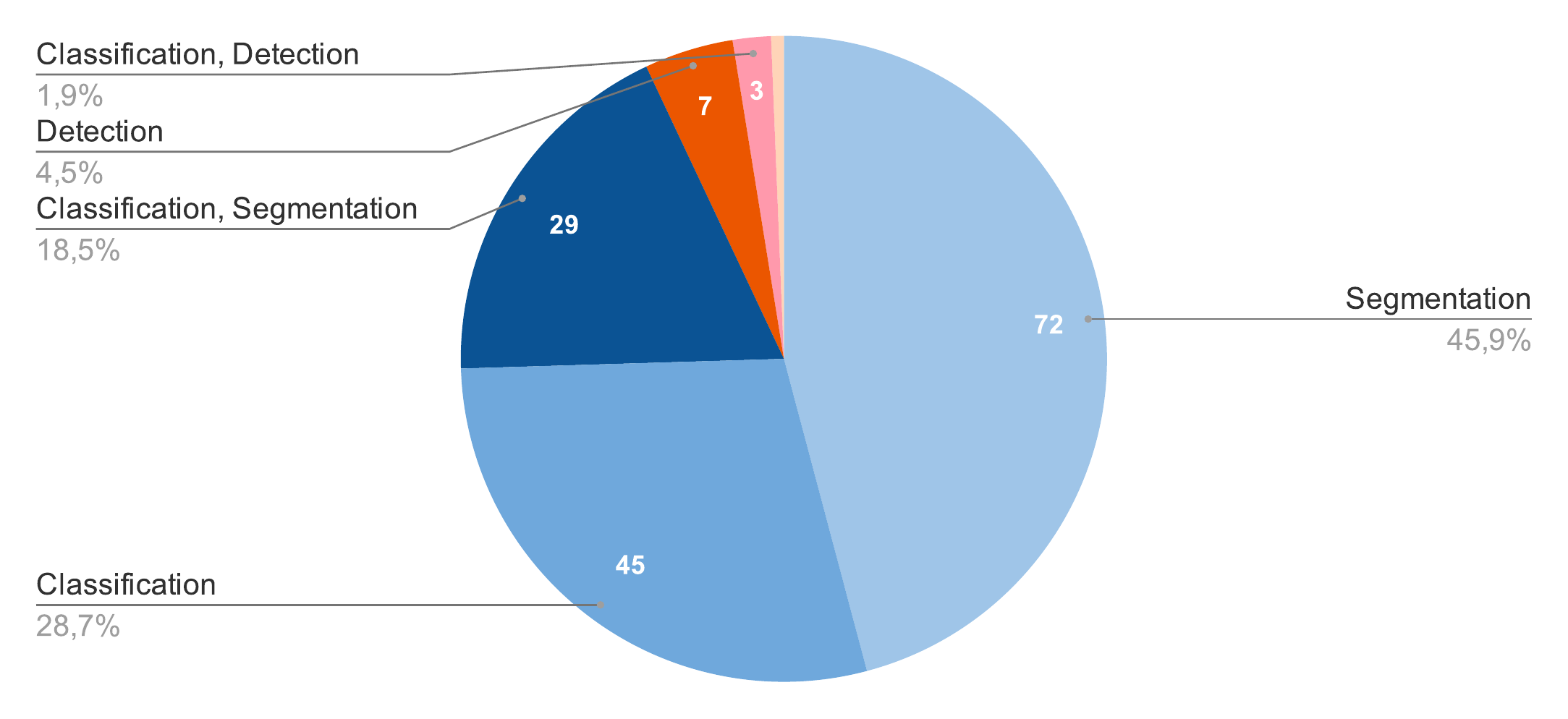}  
          \caption{Methods distribution}
          \label{graph:percertOverallMethod}
        \end{subfigure}
        \caption{Number of papers by method (Segmentation, Detection, and Classification)}
        \label{graph:graphsMethods}
    \end{figure}

\begin{figure}[htp!]
        \begin{subfigure}[b]{.5\textwidth}
          \centering
          \includegraphics[width=\linewidth]{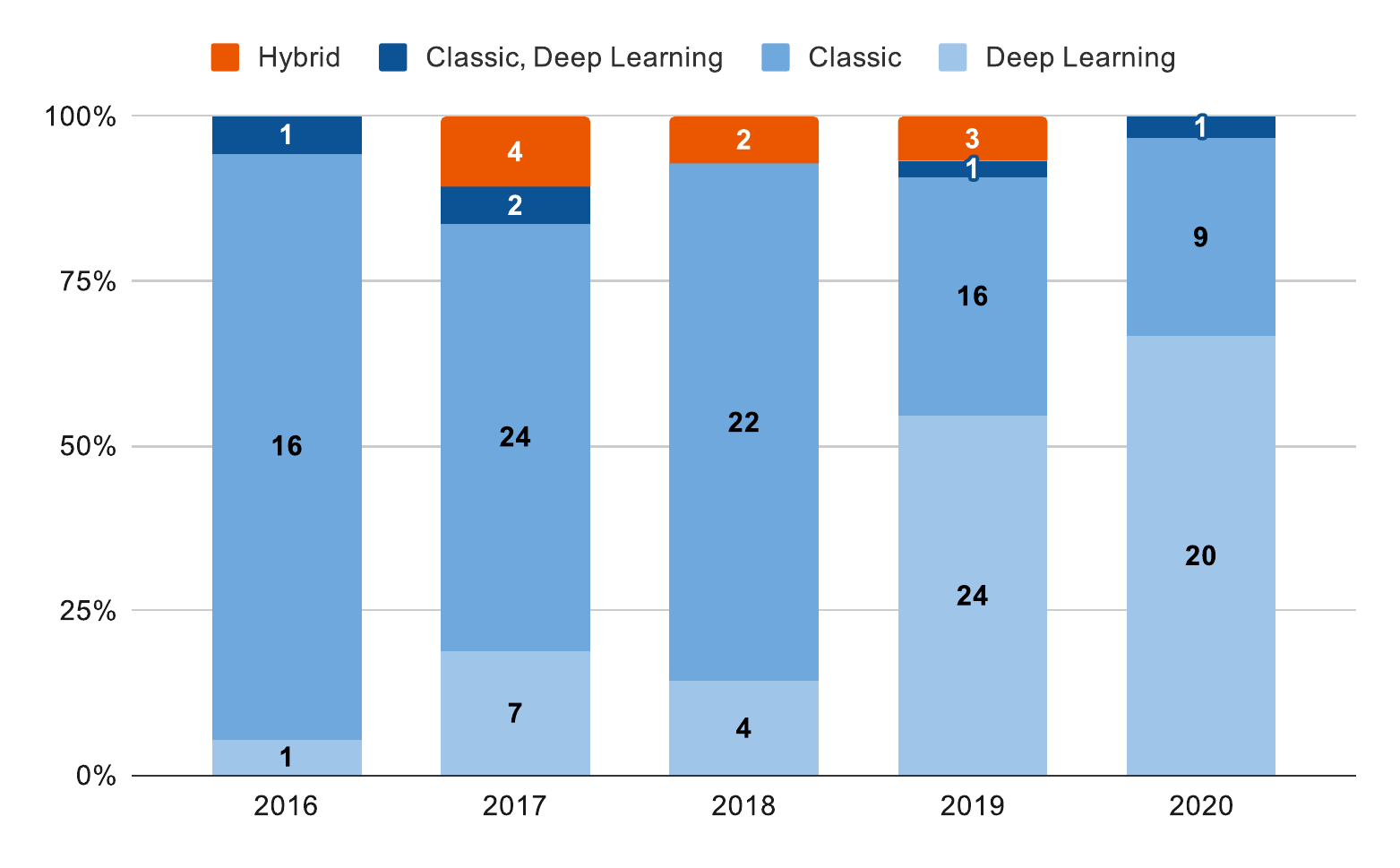}  
          \caption{Approaches by year}
          \label{graph:methodsgraph}
        \end{subfigure}\quad
        \begin{subfigure}[b]{.5\textwidth}
          \centering
          \includegraphics[width=\linewidth]{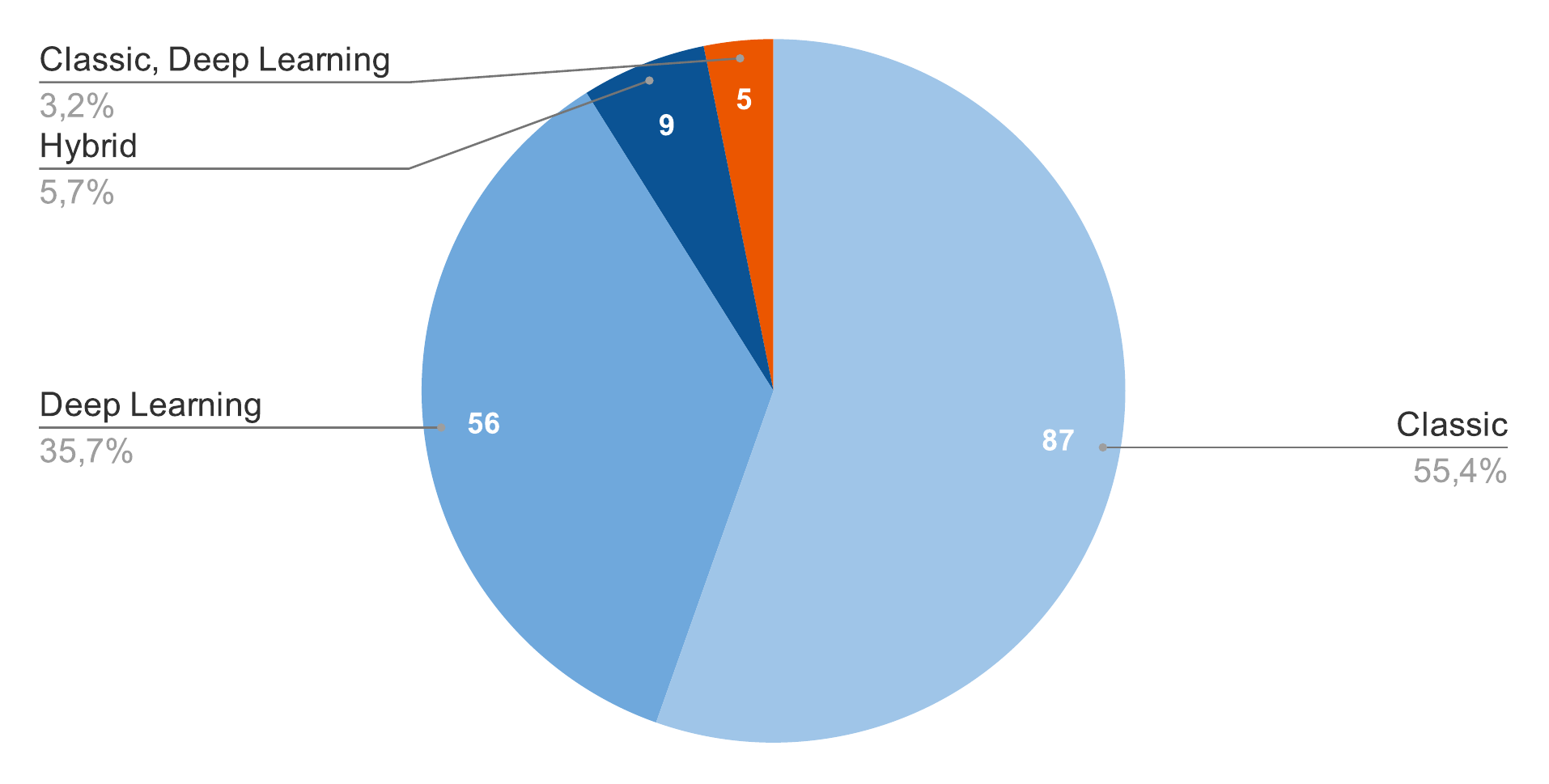}  
          \caption{Approaches distribution}
          \label{graph:percertOverall}
        \end{subfigure}
        \caption{Number of papers by approach (Deep Learning, Classic, and Hybrid)}
        \label{graph:graphWorks}
    \end{figure}

    Figure \ref{graph:graphsMethods} shows the works published by year and the categories into which we classified each work. Looking at the chart (Figure \ref{graph:percertOverallMethod}), we can see that the number of published papers on the subject of computer-aided analysis of cytology images and the variety of methods used is growing. This fact could be explained by the growing adoption of deep learning methods (Figure \ref{graph:methodsgraph}). In the last few years, various network architectures for different approaches (object detection, semantic segmentation, among others) were published for the most diverse type of applications. Some of them, i.e., U-Net, were created specifically for biomedical applications.
    
    Most of the analyzed papers use segmentation for cell localization and feature extraction  (64.4\%). Object detection is starting to be explored for those purposes, even though it is still not broadly employed (6.4\%). Besides that, Figure \ref{graph:papersbyyear} also shows a growing tendency of one-step methods while two-step methods, like the ones that use segmentation before classification, show a declining tendency over the last five years. Classification is used in 49,1\% of the papers and it is mostly applied using single-cell images. Deep learning technologies have gained a notorious space in the computer vision community in this decade, paving the way to great advances in the machine learning area. With this in mind, it is safe to assume that in the next years the number of works using \ac{CNN} technology will grow considerably. However, as shown in Figure \ref{graph:percertOverall}, classic approaches are still a relevant research area in the automated analysis of cytology slides, being used in 64.3\% of the analyzed papers. Most of the analyzed works aim to find cell abnormalities that indicate the diagnosis of cancer. The exception are 5 papers that aim to detect pleural effusions \cite{Baykal2017, Kilic2019, Baykal22019} and vulvovaginal candidiasis \cite{Momenzadeh2018,Momenzadeh2017}.
    
    The most widely employed stain was Papanicolaou, present in 130 papers, followed by H\&E stain (20 papers), Giemsa (5 papers), and Feulgen stain (5 papers).
    Pap smear datasets are used in several papers, particularly the Herlev, ISBI 2014 and 2015 challenges datasets. DCCL was the largest Papanicolaou dataset found in the analyzed works. Table \ref{tab:datasets} shows the public datasets we were able to identify. Most datasets found comprise single-cell Pap smear cytology images and the most studied pathology is cancer (cellular proliferation).

    \begin{table}[htp!]
    \scriptsize
    \centering
    \bgroup
    \def\arraystretch{1.2}
    \begin{tabularx}{\textwidth}{|>{\hsize=.115\textwidth}X|>{\hsize=.05\textwidth}X|>{\hsize=.12\textwidth}X|>{\hsize=.08\textwidth}X|>{\hsize=.1\textwidth}X|>{\hsize=.1\textwidth}X|>{\hsize=.1\textwidth}X|>{\hsize=.09\textwidth}X|}
    \hline
    
    \centering\textbf{Name} & \centering\textbf{Year} & \centering\textbf{Staining technique} & \centering\textbf{Dataset size} & \centering\textbf{Patients observed} & \centering\textbf{Anatomy} & \centering\textbf{Reference} & \textbf{Papers that have used}
    \\ \hline
    
    DTU/Herlev Pap Smear Databases & \centering{1999, 2003, 2005} & \centering{Papanicolaou} & \centering{917 cells} & \centering{-} & \centering{Cervix} & \cite{86c7f5c1a9f84a7484731dd71671c563} & \cite{Saha2018, Zhang32017, Khamparia2020, Mousser2019, Win2019, Bora2016, Indrabayu2017, Bora2017, AnousouyaDevi2018, William22019, Zhao2016, Shanthi2019, Taha2017, fekri-ershad_pap_2019, NirmalJith2018, Gautam2017, Arya2019, Zhao22019, faturrahman_multi_2017, bandyopadhyay2020segmentation, singh2018semi, zhao_automated_2019, yang_interacting_2020, ramdhani_hierarchical_2017, acharya2018segmentation, gv_automatic_2019, william_automated_2019, sabeena_improved_2020, lin2019fine, prum_abnormal_2018, saha2019prior, dumripatanachod2016centralized, lin2017llc, mishra2018theoretical, Zhang22017, pmid32308878, Wasswa2019, Somasundaram2020}
    \\ \hline
    
    HEMLBC & \centering{2014} & \centering{Liquid-based H\&E} & \centering{51 fields} & \centering{21} & \centering{Cervix} & \cite{Zhang2014} & \cite{Zhang32017, Zhang22017}
    \\ \hline

    Overlapping Cervical Cytology Image Segmentation Challenge - ISBI 2014 and 2015 & 2014, 2015 & \centering{Liquid-based Papanicolaou} & \centering{945 fields} & \centering{-} & \centering{Cervix} & \cite{ZhiLu2015, Lu2017} & \cite{Saha2018, Phoulady2017, Saha2016, Zhang2019, Wan2019, Saha22018, Saha2019, Xu2018, Arajo2019, Diniz2020, Phoulady2016, Mufidah2017, bhan2016supervised, wang_nucleus_2020, nisar2017segmentation, oliveira_multi-objective_2017, Saha2017}
    \\ \hline

    \centering{-} & \centering{2019} & \centering{Papanicolaou} & \centering{212 fields} & \centering{20} & \centering{Breast} & \cite{onedrivedataset} & \cite{Saikia2019}
    \\ \hline
    
    \centering{-} & \centering{2019} & \centering{Papanicolaou} & \centering{194 slides} & \centering{-} & \centering{Cervix} & \cite{Arajo2019} & \cite{Arajo2019}  
    \\ \hline
    
    DCCL & \centering{2019} & \centering{Liquid-based Papanicolaou} & \centering{14,432 fields} & \centering{1167} & \centering{Cervix} & \cite{Zhang22019} & \cite{Zhang22019}   
    \\ \hline
    
    SIPaKMeD & \centering{2018} & \centering{Papanicolaou} & \centering{4,049 cells} & \centering{-} & \centering{Cervix} & \cite{plissiti2018sipakmed} & \cite{plissiti2018sipakmed, gv_automatic_2019} 
    \\ \hline
    
    Perolehan Citra Kanker Serviks & \centering{2017} & \centering{Papanicolaou} & \centering{45 cells} & \centering{-} & \centering{Cervix} & \cite{Serviks:2020} & \cite{riana2017segmentation} 
    \\ \hline

    AGMC-TU & \centering{2018} & \centering{Papanicolaou} & \centering{225 cells} & \centering{45} & \centering{Cervix} & \cite{Bhowmik2018} & \cite{Bhowmik2018} 
    \\ \hline
    
    CCAgT & \centering{2020} & \centering{AgNOR} & \centering{2,540 fields} & \centering{3} & \centering{Cervix} & \cite{AtkinsonAmorim2020, CCAgT2020} & \cite{AtkinsonAmorim2020} 
    \\ \hline
    
    UFSC-OCPap & \centering{2020} & \centering{Papanicolaou} & \centering{1,934 fields} & \centering{2} & \centering{Oral} & \cite{Matias2020, papmatias2020} & \cite{Matias2020} 
    \\ \hline
    
    \end{tabularx}
    \egroup
    \caption{Public datasets}
    \label{tab:datasets}
\end{table}

\begin{figure}[H]
    \begin{center}
    	\includegraphics[width=\linewidth]{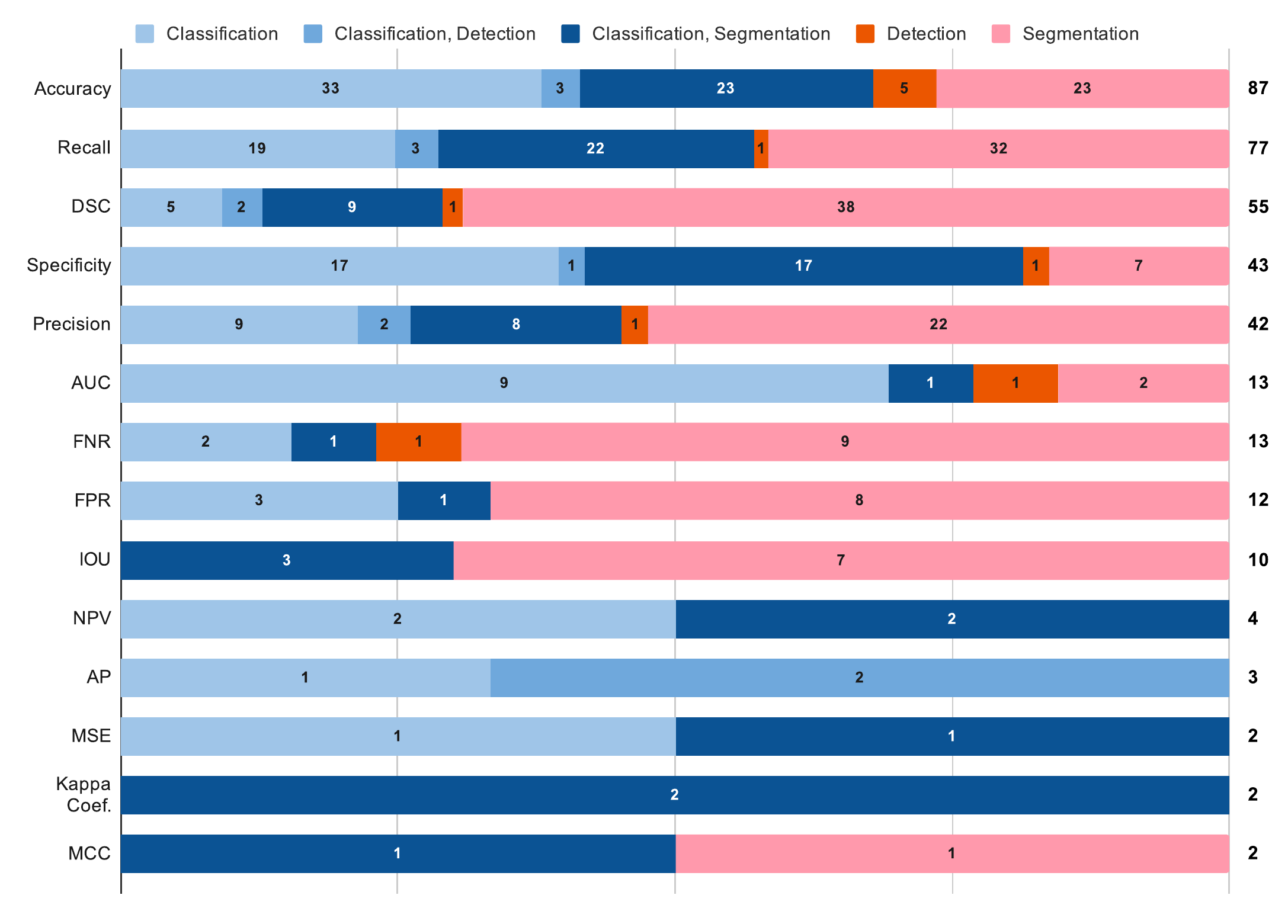}
    	\caption{Evaluation metrics distribution by approach}
    	\label{graph:metrics}
    \end{center}
\end{figure}

We categorize the uses of each metric based on the type of approach the paper uses, and a supplementary analysis is available in \ref{appEva} and \ref{tab:metricsPapers}. Summarizing here an overview of the most used metrics, we identified that the most recurrent metrics used were \emph{Accuracy}, which appears in 87 papers, and \emph{Recall}, found in 77 papers. The third most used and the dominant metric in segmentation approaches was the \emph{Dice Similarity Coefficient}, which was used in 55 papers, with 38 of them using segmentation approaches. The \emph{Specificity} was the fourth most used metric, as shown in Figure \ref{graph:metrics}. Even though these four metrics are the most-used, there is no common sense, since in this \ac{SLR} we could identify 21 different evaluation metrics. 

We also identified the most commonly used training strategies, shown in Table \ref{tab:training_approach} and the supplementary analysis available in \ref{apptrainap}. The train-test scheme was the most used in the analyzed papers (64 papers). Of these, 41 are in articles that use ML and 23 that use classic approaches. We also point out that classics approaches tend not to use cross-validation techniques, as 25 CV papers and 17 ML papers use the full dataset for training and test.

\begin{table}[H]
    \centering
    \begin{tabular}{|c|c|c|}
    \hline 
    \multirow{2}{*}{\textbf{Training Approach}} & \multicolumn{2}{c|}{\textbf{Number of Papers}} \\ \cline{2-3} 
                                      & \textit{ML}        & \textit{Classic}        \\\hline 
                    Train-Test         & 41                 & 23\\\hline 
                    $k$-Folds             & 24                 & 7\\\hline 
                    TVT                & 21                  & 0\\\hline 
                    Full Dataset       & 16                  & 25\\\hline 
    \end{tabular}
    \caption{Training and dataset handling approaches}
    \label{tab:training_approach}
\end{table}

\section{Discussion}

We observed a repetition of techniques and pipelines in groups of articles, with minor variations in the processing techniques applied. Many of these groups use similar techniques, and this led us to group them into large groups. We tried, when it made sense, to analyze the nuances and differences of each group. However, in some cases, a differentiation between papers did not make sense and that is why some works are cited only inside groups. Mostly, these groupings occurs with popular models like ANNs, SVM and Decision Trees when they are applied without great modifications.

The yearly distribution by CV approach of the works, as shown in Figure \ref{graph:graphWorks}, indicates a crescent preponderance of DL-based methods. This is expected because the recent advances in image processing, especially in the field of modern Deep Learning techniques, have been very successful in addressing previously unsolvable problems in image detection, classification, and semantic segmentation, in different areas of application. Problems that resisted previous attempts of researchers in the field of machine learning and artificial intelligence were solved using \ac{CNN}. The results achieved by CNN techniques exceed the previous ones by a large margin \cite{LeCun2015}. This way, it is natural that the researchers will try to employ these approaches to solve problems in the cytology field as well.

One of the most significant drawbacks of DL methods is the need for numerous samples to train a model. Even though we could identify some public datasets, sometimes it is necessary to gather specific data, from which there is no public dataset available, making it necessary to spend a large amount of time as a specialist to label a large amount of data to train a model. However, it is important to consider that CNNs work like a general-purpose feature extractor by adjusting the network weights using the error-backpropagation algorithm, generating complex filters that can be used to detect complex patterns \cite{krizhevsky2012}. Considering all the articles reviewed, about 35.7\% use deep learning, as shown in Figure \ref{graph:graphsMethods}, but considering the data from the last 2 years, about 59.46\% of the papers use deep learning. 

Most of the networks identified in this SLR are an application of unmodified architectures that appear in the literature (VGG-16, VGG-19, Inception V3, ResNet50, ResNet101, among others). There are only two novel architectures, the W-Net (double U-Net) and the TernausNet, composed of a modified U-Net and the DeepLabV2. For semantic segmentation, the U-Net was the most recurrent architecture. Based on this information, we could infer that the researchers are most concerned with solving the problem than create/develop new CNN architectures. In other words, this problem was not solved yet. According to this and by taking an overview look at the works found on this SLR, we could notice that most of the approaches are not ready yet to be applied on a clinical routine. 

On the other side,  traditional CV approaches are much more specific and highly parameter-dependent. This can explain why classical approaches found in the literature showed to be less effective when compared with DL models \cite{kamilaris2018deep}. Classic CV techniques are strongly dependent on algorithm parameters and also on image characteristics such as contrast, noise, and resolution, making them much less robust than more state-of-the-art CNN-based approaches \cite{omahony2019}. This suggests that \ac{CNN}s  are better suited for the application of computer-aided diagnosis for cytology images. Moreover, developing a hand-engineered feature extractor to use classical approaches is a difficult trial-and-error process, that involves not only mathematical knowledge about CV algorithms, but also domain knowledge that takes into account the nature and complexity of the problem. Besides that, traditional CV still seems to be an active field of research for cytology analysis applications, considering that classic and hybrid approaches are used in 64.3\% of the selected papers, as shown in Figure \ref{graph:graphWorks}.

 Most of the analyzed works aim to find cell abnormalities that indicate the diagnosis of cancer. By taking into account that 152 of 157 papers presents methods to detect cancer, this seems to be the main concern of the researchers. It could be explained by the fact that some cancer types are curable if early detected. An example is the cervical cancer. It is a curable type of cancer if it could be early detected and treated properly \cite{song2016accurate}. The exception are the papers that focus on lung cells. This type of works have the goal of detect pleural effusions \cite{Baykal2017, Kilic2019, Baykal22019} and vulvovaginal candidiasis \cite{Momenzadeh2018, Momenzadeh2017}. Figure \ref{graph:graphWorks} shows the number of each type of approach found in the reviewed works. The ``Classic" category stands for classical image processing methods. The category ``Deep Learning" represents any deep learning-related approach, and ``Hybrid" accounts all the works that used both of method types in the same pipeline. The ``Classic, Deep Learning" category represents the papers that test both methods in distinct pipelines.

The most widely employed stain was Papanicolaou, present in 130 papers. However, it is necessary to point out some facts. There is a lack of standardization on its composition, coming in several formulations. The dyes that are used, their ratios, and the timing of the process could be different on different stains \cite{Gill2013, schulte1991standardization}. Some works present evidence that the analysis of cytological material using the Papanicolaou staining is not a reliable method for DNA quantification \cite{gurley1990comparison}. A study investigating the suitability of Papanicolaou staining and hematoxylin staining in comparison with Feulgen staining for DNA single-cell cytometry found that histograms of Feulgen stained normal squamous epithelia showed a regular DNA distribution, while Papanicolaou and hematoxylin staining showed a wide scatter of values and an increased number of values above 4c. Thus, DNA measurements from Papanicolaou-stained and hematoxylin-stained specimens are not suitable to draw diagnostic or prognostic conclusions because they cannot be interpreted reliably due to the lack of stoichiometry of nuclear staining \cite{biesterfeld2011feulgen}.

We identified the size of the datasets and the imbalance of classes as the most recurrent problems in the analyzed papers. This indicates that there is still a lack of large and qualitatively reliable datasets comprising samples from real-world cytology examinations, as most of the approaches are based upon controlled samples (ISBI 2014/2015) or pre-processed images (Herlev). Given this, we considered DCCL as the most promising dataset and we believe it should be more explored as it was used only on its own benchmark work. It employs liquid-based cytology, as most of the datasets that we found do, which is less cost-effective than conventional cytology. We understand that this points out the necessity of good-quality datasets with images gained from conventional cytology examinations. Also, there is a lack of datasets composed of less used stains, such as the Feulgen, for example. The availability of datasets with different stains could present a direction for the researcher on defining which stain presents the best results for the attended objective.

The differences in validation strategies and metrics indicate that a simple and direct comparison between papers is not always fair because some of them, aside from distinct evaluation metrics, also use private or new datasets. One method for the measurement of the performance of object detection and semantic segmentation models that has become popular recently in the CNN literature is the Intersection Over Union (IoU) \cite{ZHAO2019}, as shown in Figure \ref{figure:graphmetrics}. Cytology is an ideal field of application for the IoU because objects are all of similar size and the \emph{large object bias} problem does not occur \cite{visapp18}. IoU is also considered not to be so much affected by distribution bias problems, as is the case of accuracy and similar metrics.
In the cytology field, however, IoU was employed in only ten of the works we analyzed. However, it is important to emphasize the importance of using cross-validation, especially when working with medical/cytology data, to avoid a bias in its results.

To validate our SLR strategy to avoid evidence selection bias of also searching the scientific publishers' databases directly and not only on metasearch engines such as PubMed or NCBI, we retrospectively searched on PubMed and NCBI explicitly for papers that our direct search on the scientific publishers' databases identified as relevant and, e.g., \cite{Arya2016, Sudheesh2016, Toutain2016, Phoulady2016, Nawandhar2020} are relevant publications that could not be found on neither PubMed nor NCBI. Most papers we did not find on PubMed or NCBI were ACM- or IEEE-published papers, published in CS or engineering journals, but even \cite{Nawandhar2020}, which was published in the \emph{Journal for Biomedical Signal Processing and Control}, an Elsevier journal available on Science Direct, could not be found on either PubMed or NCBI. We considered that these results corroborated our strategy.

\section{Conclusions}

    Analyzing the papers collected in this review, we could determine that there is no publicly available dataset for many of the stains used in the described experiments. Most of the papers employ the Liquid-based Papanicolaou, even though there are variations on the stain. 
    One impression is that many of these studies are still in a very experimental phase and not ready to be applied on a daily clinical routine. This is expected, since most of them are adopting \ac{DL} methods, and this technology is relatively new, even though it advances exponentially. Looking at these works and comparing to the general tendencies that are observable in the CV community in general, we can see that, in the next few years, there will probably be a still larger adoption of \ac{DL} methods for studies in the cytology image processing area, although classic Computer Vision methods are still being explored as solutions to the automation of cytology analysis.
    
\subsection{Threats to Validity}

This SLR was explicitly formulated in order to take into consideration only cytology papers. There exists a much vaster literature on CV for histopathology that we did not consider here. From the point of view of our knowledge on CV, we understand that many of the CV techniques used in histopathology could also be applied without much modification to cytological samples.  This was a conscious delimitation of this study in order to focus on the cytology field. 

\section{Acknowledgements}

This study was financed in part by the Coordenação de Aperfeiçoamento de Pessoal de Nível Superior - Brasil (CAPES), the Brazilian Federal Agency for Support and Evaluation of Graduate Education and by the FAPESC - Santa Catarina State Research Support Foundation. It was also supported by the Brazilian National Institute for Digital Convergence (INCoD), a research unit of the Brazilian National Institutes for Science and Technology Program (INCT) of the Brazilian National Council for Science and Technology (CNPq).

\newpage
\appendix
\section{Table with infos of all papers}
\begin{tiny}
\label{tab:reviwed_works}
\begin{longtable}[c]{|c|c|c|c|c|c|}
    \hline
    \textbf{paper}               & \textbf{year} & \textbf{approach}            & \textbf{method}        & \textbf{stain} & \textbf{dataset}               \\ \hline
    \endfirsthead

    \multicolumn{6}{c}
    {{\bfseries Table \thetable\ continued from previous page}} \\
    \hline
    \textbf{paper}               & \textbf{year} & \textbf{approach}            & \textbf{method}        & \textbf{stain} & \textbf{dataset}               \\ \hline
    \endhead

    \cite{GarciaGonzalez2016}             & 2016 & \begin{tabular}[c]{@{}c@{}}Classification, \\ Segmentation\end{tabular} & Classic                & Pap             &                         \\ \hline
    \cite{Phoulady2016}                   & 2016 & Segmentation                 & \begin{tabular}[c]{@{}c@{}}Classic and\\Deep Learning\end{tabular} & Pap             & ISBI 2015               \\ \hline
    \cite{Toutain2016}                    & 2016 & \begin{tabular}[c]{@{}c@{}}Classification, \\ Segmentation\end{tabular} & Classic                & \begin{tabular}[c]{@{}c@{}}Pap,\\Feulgen\end{tabular}     &                         \\ \hline
    \cite{fcheng-2016}                    & 2016 & Segmentation                 & Classic                & Pap             &                         \\ \hline
    \cite{Zhao2016}                       & 2016 & \begin{tabular}[c]{@{}c@{}}Classification, \\ Segmentation\end{tabular} & Classic                & Pap             & Herlev                  \\ \hline
    \cite{Su2016}                         & 2016 & \begin{tabular}[c]{@{}c@{}}Classification, \\ Segmentation\end{tabular} & Classic                & Feulgen         &                         \\ \hline
    \cite{neghina_automatic_2016}         & 2016 & Segmentation                 & Classic                & Pap             &                         \\ \hline
    \cite{dumripatanachod2016centralized} & 2016 & Classification               & Classic                & Pap             & Herlev                  \\
\end{longtable}
\end{tiny}

\section{Table with details of classical segmentation papers}
\label{tab:papers-segmentation-classic}
\begin{tiny}
\begin{longtable}[c]{|>{\raggedright}m{0.8cm}|>{\raggedright}m{2.3cm}|>{\raggedright}m{3.7cm}|>{\raggedright}m{2.6cm}|m{2.1cm}|}
\hline
\textbf{Paper}                                                               & \textbf{Objective}                             & \textbf{Pre/Pos-processing}                                       & \textbf{Classification}                                                           & \textbf{Segmentation}                    \\
\hline
\endfirsthead

    \multicolumn{5}{c}
    {{\bfseries Table \thetable\ continued from previous page}} \\
    \hline
    \textbf{Paper}                                                               & \textbf{Objective}                             & \textbf{Pre/Pos-processing}                                       & \textbf{Classification}                                                           & \textbf{Segmentation}                \\ \hline
    \endhead

\cite{oliveira_multi-objective_2017}                                         & Detect/Count Cells                                 & -                                                                     & -                                                                                 & Clustering (DBSCAN, \ac{SLIC} superpixel merging) \\
\hline
\cite{wang_nucleus_2020}                                                     & Detect/Count Cells                                 & Noise reduction                                                       & -                                                                                 & Thresholding                                                  \\
\hline
\cite{neghina_automatic_2016}                                                & Detect/Count Cells                                 & Geometric shape constraints, iso-contours                             & -                                                                           & Clustering (k-means)                                                             \\
\hline
\cite{Win2017}                                                               & Detect/Count Cells                                 & Conversion to l*a*b*, noise reduction, morphological operations       & -                                         & Thresholding                                                  \\
\hline
\cite{Baykal2017}                                                            & Detect/Count Cells                                 & Dimension reduction                                                   & -                                                                   &  Trainable Model (Active appearance model)                                                            \\
\end{longtable}
\end{tiny}

\section{Training and Validation Strategies}\label{apptrainap}

This section describes the different strategies we identified that were used to handle the datasets in order to train and test the chosen \ac{ML} algorithms, and the validation strategies and methods used to validate the results.

\subsection{Train-Test Split and TVT Model}

In order to avoid biased results, the dataset is commonly split into two disjunct subsets: \emph{train} and \emph{validation} set. The \textit{train set} is used to train the \ac{ML} algorithm and the \textit{validation set} is used to validate the results, demonstrating the generalization capability of the trained model.

Sometimes the validation set is also used for parameter or hyperparameter tuning during training: network performance is validated at each epoch and validation results are used to optimize the training process, influencing hyperparameters such as learning rate or momentum. In this case, the validation set also has an influence on the training process, which can result in overfitting. In order to avoid that, a more robust approach is the utilization of the \ac{TVT} Model,
where the full dataset is split into three disjunct subsets \cite{ripley2007pattern}, as shown in Figure \ref{diagram:tvt}:

    \begin{figure}[H]
        \begin{center}
        	\includegraphics[width=1.0\linewidth]{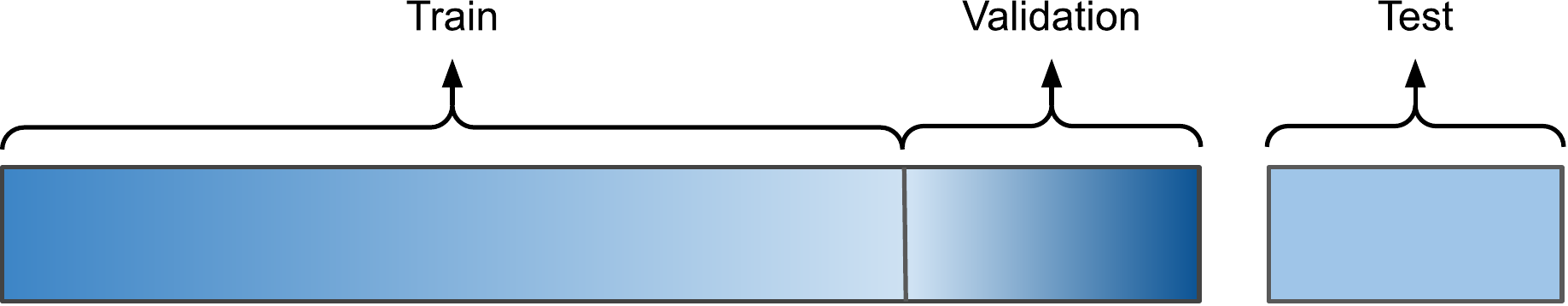}
        \end{center}
        \caption{A visual representation of the TVT Model}
        \label{diagram:tvt}
    \end{figure}

\begin{itemize}
    \item Training set: The subset used for training the model;
    \item Validation set: The subset used to test the network during training and for parameter or hyperparameter tuning;
    \item Test set: The subset used only after training and parameter tuning to assess the performance of a fully-specified model.
\end{itemize}

The terms ``validation set" and ``test set" are used interchangeably in some of the papers, so we used the definitions above to perform the analysis. Also, there are 9 papers that used k-Fold for parameter tuning, instead of using distinct sets for train and validation, and these were not classified as using the \ac{TVT} Model.

\subsection{k-Folds}
$k$-folds cross-validation is a validation method where, for every $k$ iteration, the dataset is randomly partitioned into $k$ subsets of equal size. Then, one of these $k$ subsets is used to validate the model while the remaining $k-1$ are used for training.

    \begin{figure}[htp!]
        \begin{center}
        	\includegraphics[width=\linewidth]{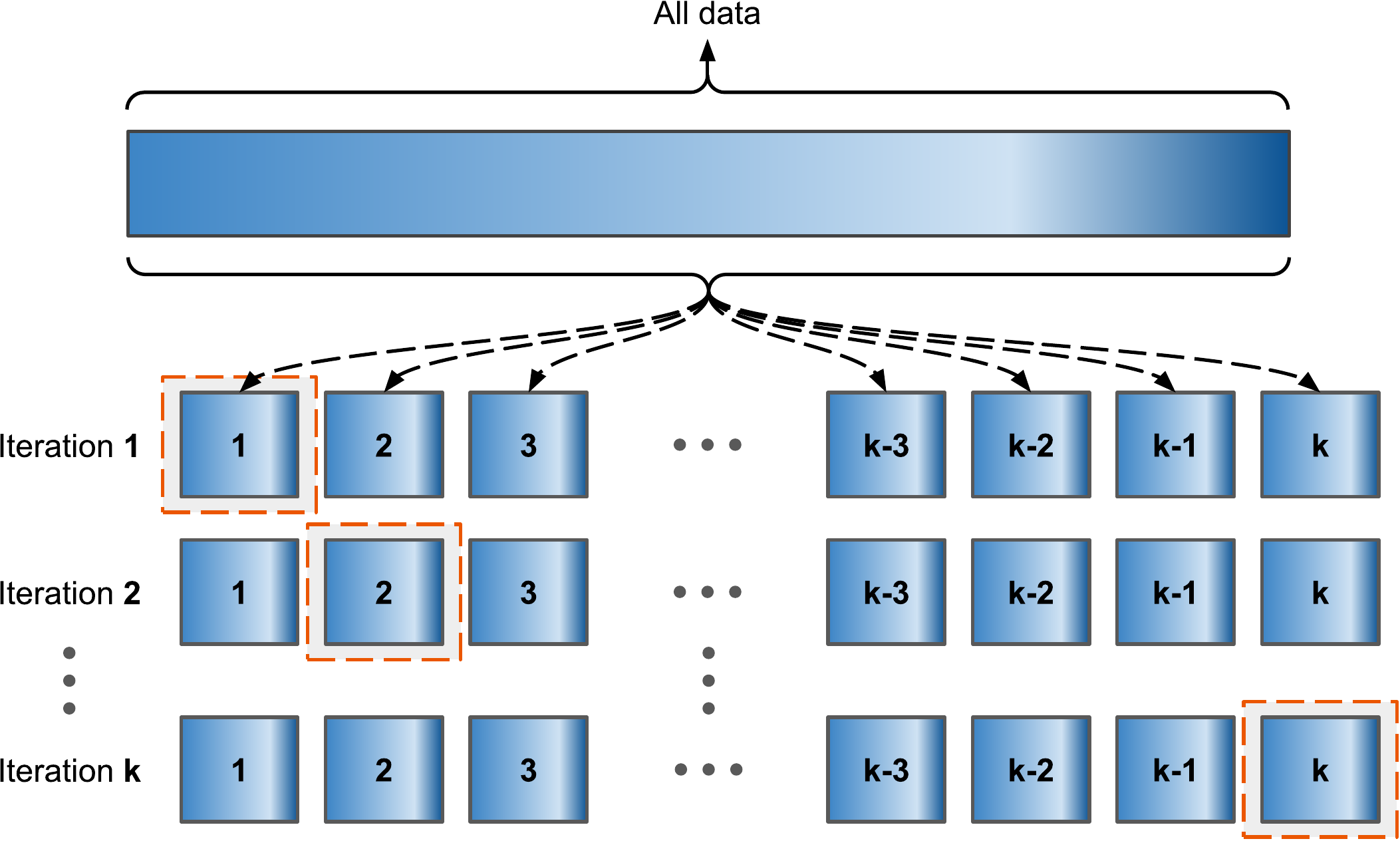}
        	\caption{A visual representation of the k-Folds cross validation}
        	\label{diagram:kfold}
        \end{center}
    \end{figure}

This folding process is repeated $k$ times, resulting in each subset being used exactly one time for validation. The final result is the average of the $k$ results obtained. The main advantage of this method is  that the entire dataset is used for training and validation and each sample is used for validation exactly one time \cite{mclachlan2005analyzing}. Figure \ref{diagram:kfold} shows a visual example of the $k$-folds cross-validation.

\section{Evaluation Metrics}\label{appEva}

We have identified several different validation strategies and metrics that were employed in order to evaluate the approaches described in the papers. Only metrics that are used in more than one of the analyzed papers are explained.
Furthermore, in the context of multi-class  detection, there are many classes, which may not be uniformly distributed. This would indicate, for example, that an accuracy-based metric may introduce biases \cite{SOKOLOVA2009, LIU2020}. 
    
    The following metrics were not included  due to having been used only once throughout our review: Piccard Index \cite{Khamparia2020}, Support \cite{Khamparia2020}, Relative Detection Error (RDE) \cite{Bhowmik2018}, Aggregated Jaccard Index \cite{Bhowmik2018}, Hausdorff distance \cite{Bhowmik2018}, PQ \cite{Bhowmik2018}, Spearman rank-order correlation coefficient \cite{Bhowmik2018} and Correlation Coefficient \cite{Bhowmik2018}. There is also the Mean Square Error (MSE) used at \cite{dumripatanachod2016centralized, Bhowmik2018, Matias2020}.
    
    In order to describe each metric, it is first necessary to describe the four basic parameters used to calculate these metrics. These parameters are listed below:
    \begin{itemize}
        \item \textbf{True Positive (TP)}: also known as ``hits"; number of correct positive predicted values. 

        \item \textbf{True Negative (TN)}: also known as the ``correct rejection"; is when the value is false and the prediction is equal the ground truth. 
        
        \item \textbf{False Positive (FP)}: erroneous detection between of the prediction value and the ground truth. 
        
        \item \textbf{False Negative (FN)}: ground truth not detected; in other words, the ``misses".
    \end{itemize}
    
    The notation applied in this work for TP, TN, FP, and FN for pixel-wise metrics is showed in the Table \ref{tab:metricsEq}, where: $n_{jj}$ in TP is the number of pixels both classified and labelled as class $j$. Similarly, $n_{ff}$ in TN is the number of pixels both classified and labelled as class $f$, the correct rejection. For false cases, $n_{ij}$ in FP is the number of pixels that are labelled as $i$ class, but classified as $j$. Finally, $n_{ji}$ in FN is the number of pixels that are labelled as $j$ class, but classified as $i$ \cite{Fawcett2006, powers2011evaluation}.

\begin{figure}[htp!]
        \begin{center}
        	\includegraphics[width=1.0\linewidth]{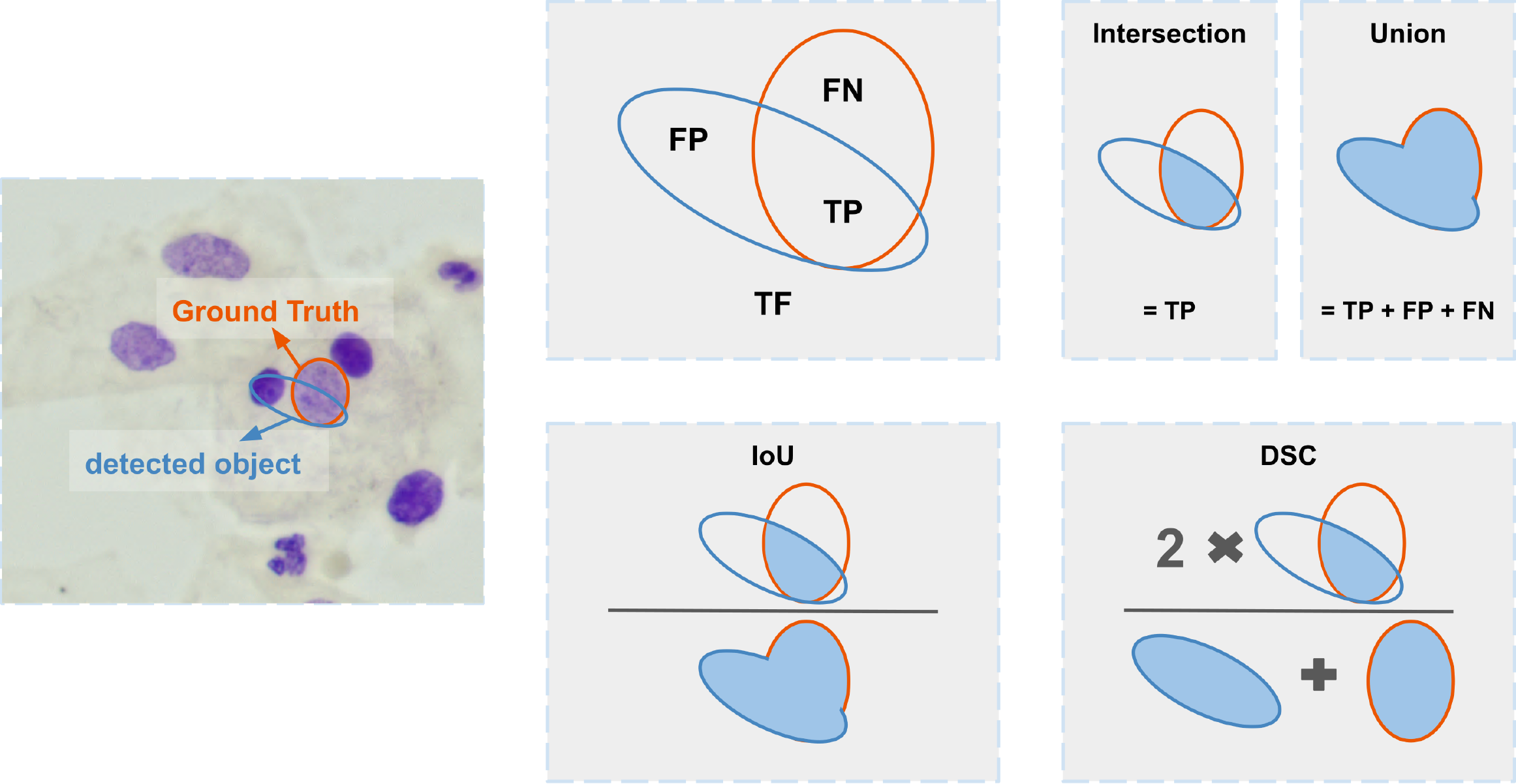}
        	\caption{Example of the application of the metrics}
        	\label{figure:graphmetrics}
        \end{center}
    \end{figure}

    These parameters can be used for classification, object detection, and semantic segmentation changing its form of implementation, and consequently changing the metrics for each desired method. For classification and object detection the implementation is similar, they work directly with the quantity of objects/items detected/classified. For semantic segmentation the implementation needs to be based on the number of pixels for each class. The equations presented in Table \ref{tab:metricsEq} show both approaches, where all equations are based on these four parameters, which can also be used as a sum of those pixel occurring, leading to a quantification of the number of pixels. This is shown visually in the Figure \ref{figure:graphmetrics}, where we explain what are the union and intersection cases in these terms.
    
    For better comprehension the parameters (TP, TN, FP and FN) for \textbf{object detection} are explained below:
    
    \begin{itemize}
        \item \textbf{True Positive (TP)}: It is considered a TP when the IoU between the predicted object and the ground truth is higher than a given threshold (usually $IoU>0.5$)

        \item \textbf{True Negative (TN)}: TN is not used in object detection because it usually makes no sense to train a model with a negative case, or ``background", object
        
        \item \textbf{False Positive (FP)}: When the model detects an object that is not present in the ground truth or it is present but the IoU is below a given threshold (usually 0.5).
        
        \item \textbf{False Negative (FN)}: When the object is present in the ground truth and the model does not detect it.
    \end{itemize}
    
    Also, those parameters need some extra considerations: Only the prediction with the highest IoU is considered TP for one object, the others are FP; If there is an object in the ground truth, and the model predicts it but with an IoU below the threshold, it must be accounted as FP, but if there are no predictions for this object, it must be accounted for as FN; If a prediction has no corresponding ground truth, it is considered as FP.

    In a similar way, the metrics can be written for both, binary- or multi-class approaches. In papers where the authors use multi-class approaches, the class-averaged metric value is usually used. However, this can be computed using also  frequency-weighted values or the value for each class \cite{SOKOLOVA2009, ulku2019survey}.

    The explanation for each metric presented in Figure \ref{graph:metrics} is enumerated below (1 to 13), and its equations are shown in the Table \ref{tab:metricsEq}. The \textit{Support} metric is not listed because it represents the number of occurrences of each class in the GT.

\begin{enumerate}
    \item \textbf{Precision}: is the ratio between the \textit{True Positives} and the sum of \textit{False Positives} and \textit{True Positives} \cite{Fawcett2006, powers2011evaluation}. This metric is also called Positive Predictive Value (PPV) and shows how accurate is the model.

    \item \textbf{Recall or Sensitivity}: is the ratio between the \textit{True Positives} and the sum of \textit{False Negatives} and \textit{True Positives} \cite{Fawcett2006, powers2011evaluation}, this equation is also known as of True Positive Rate (TPR). This metric shows the probability to classify the positive cases correctly.

    \item \textbf{Negative Predictive Value (NPV)}: is the ratio between the \textit{True Negatives} and the sum of \textit{False Negatives} and \textit{True Negatives} \cite{powers2011evaluation}. This equation also known as True Negative Accuracy (TNA) and indicates the probability to classify negative instances correctly.

    \item \textbf{False Positive Rate (FPR)}: is the ratio between the \textit{False Positives} and the sum of \textit{False Positives} and \textit{True Negatives} \cite{Fawcett2006, powers2011evaluation}. In other words, it is the probability of the model to make a false positive prediction.

    \item \textbf{Miss rate or False Negative Rate (FNR)}: is the ratio between the \textit{False Negative} and the sum of \textit{False Negative} and \textit{True Positive} \cite{Fawcett2006, powers2011evaluation}. This is the probability of having a negative result with a positive item present in the context.

    \item \textbf{Specificity}: is the ratio between the \textit{True Negatives} and the sum of \textit{False Positive} and \textit{True Negatives} \cite{Fawcett2006, powers2011evaluation}. This metric shows the probability of the model to classify the negative cases correctly.

    \item \textbf{Accuracy}: is the ratio between the \textit{True Negatives} and the sum of \textit{False Negatives} and \textit{True Negatives} \cite{Fawcett2006}. It indicates the overall effectiveness of the model.

    \item \textbf{Area under the ROC curve (AUC)}: the ROC curve is plotted with TPR in the y-axis and FPR in the x-axis \cite{powers2011evaluation}. The ROC curve is a probability curve, so an AUC near 1 is a good measure of separability \cite{Fawcett2006, SOKOLOVA2009}. This metric also measures the ability of a model to avoid false classifications.

    \item \textbf{Kappa Coefficient}: is a statistic used to measure the reliability or agreement between participants for categorical items (multi-classes) \cite{McHugh2012}. In its equation the $P_0$ is the $Accuracy$, or the total agreement probability, and $P_c$ is the agreement probability which is due to chance \cite{BenDavid2008}. This metric ranges from $-1$ to $+1$ and measure the inter-rater and intra-rater reliability for the classes.

    \item \textbf{Matthews Correlation Coefficient (MCC)}: A dataset imbalance measure that is usually used in binary classification problems. The closer to 1, the more balanced is the dataset \cite{Matthews1975}. 
    
    \item \textbf{Average Precision (AP)}: this metric is used mostly for object detection and is defined by the area under the Precision-Recall curve, where the precision and recall are calculated for each image. This curve is drawn with precision on the y-axis and recall on the x-axis \cite{ulku2019survey}. \cite{Everingham2009, lin2014microsoft} describe this metric in more detail. 
    
    \item \textbf{Dice Similarity Coefficients (DSC) or F-Score}: this is a metric for semantic segmentation  ans is a normalized measure of similarity, consisting of the normalized union of classes \cite{Crum2006}. This measure is the harmonic mean of the precision and recall for a given threshold \cite{ulku2019survey, powers2011evaluation} and is shown visually in the figure \ref{figure:graphmetrics}.

     \item \textbf{Intersection Over Union (IoU)}: this  is a metric for semantic segmentation and a measure based on the Jaccard Index \cite{Jaccard} that evaluates the overlap between two regions in the image: the \emph{ground truth region} and the \emph{predicted region} \cite{ZHAO2019, ulku2019survey}. A visual representation of this metric is shown in Figure \ref{figure:graphmetrics}. For pixel-wise approaches, this measure is the intersection of the pixel-wise classification results with the ground truth, to their union and this is the way used to measure the accuracy in model of detection objects or semantic segmentation.

\end{enumerate}

\begin{table}[H]
    \centering
    \bgroup
    \def\arraystretch{1.7}
    \begin{tabularx}{\textwidth}{|>{\hsize=.15\textwidth}X|X|}
    \hline
        \multicolumn{1}{|c|}{\textbf{Metric}}  & \multicolumn{1}{c|}{\textbf{Formula}} \\ \hline
        TP & $\sum_{j=1}^{k}{n_{jj}}$  \\ \hline
        TN & $\sum_{f=1}^{k}{n_{ff}}$ \\ \hline
        FP & $\sum_{j=1}^{k}{n_{ij}}, \quad i\neq j$ \\ \hline
        FN & $\sum_{j=1}^{k}{{n_{ji}}}, \quad i\neq j$ \\ \hline
        Precision & $\frac{TP}{FP + TP} = \sum_{j=1}^{k}{\frac{n_{jj}}{n_{ij} + n_{jj}}}, \quad i\neq j$ \\ \hline
        Recall or Sensitivity & $\frac{TP}{FN + TP} = \sum_{j=1}^{k}{\frac{n_{jj}}{n_{ji} + n_{jj}}},        \quad i\neq j$ \\ \hline

        FPR & $\frac{FP}{FP + TN} = \sum_{j,f=1}^{k}{\frac{n_{ij}}{n_{ij} + n_{ff}}}, \quad i\neq j, i\neq f$ \\ \hline
        FNR & $\frac{FN}{FN + TP} = \sum_{j=1}^{k}{\frac{n_{ji}}{n_{ji} + n_{jj}}}, \quad i\neq j$ \\ \hline
        Specificity &  $\frac{TN}{FP + TN} = 1 - FPR =  \sum_{j,f=1}^{k}{\frac{n_{ff}}{n_{ij} + n_{ff}}}, \quad i\neq j, i\neq f$ \\ \hline
        Accuracy & $\frac{TP + TN }{FP + FN + TP + TN} = \sum_{j,f=1}^{k}{\frac{n_{ff} + n_{jj}}{n_{jj} + n_{ff} + n_{ji} + n_{ij}}}, \quad i\neq j, i\neq f$ \\ \hline
        AUC &  $\frac{ Recall + Specificity }{2} = \frac{1}{2} \times \bigg(\frac{TP}{TP+FN}+\frac{TN}{TN+FP}\bigg)$ \\ \hline
        MCC &  $\frac{TP\times TN - FP \times FN}{\sqrt{(TP+FP)(TP+FN)(TN+FP)(TN+FN)}}$ \\ \hline
        Kappa Coefficient &  $\frac{P_0 - P_c}{1 - P_c} = \frac{TP - FP - (TP+FP)(1-2(FP+TN))}{ 1 - (TP+FP)(1-2(FP+TN)) - FP - TN}$ \\ \hline
        DSC &  $\frac{ 2 \times Precision \times Recall}{Precision + Recall} = {\frac{2 \times TP}{FP+FN+2 TP}} = \sum_{j=1}^{k}{\frac{ 2 \times n_{jj}}{n_{ij}+n_{ji}+2 n_{jj}}}, \quad i\neq j$ \\ \hline
        IoU & $\frac{TP}{FP+FN+TP} = \sum_{j=1}^{k}{\frac{n_{jj}}{n_{ij}+n_{ji}+n_{jj}}}, \quad i\neq j$ \\ \hline
    \end{tabularx}
    \egroup
    \caption{Metrics equations}
    \label{tab:metricsEq}
\end{table}

\section{Table with papers that used each metric}
\label{tab:metricsPapers}
    \begin{table}[H]
    \centering
    \bgroup
    \def\arraystretch{1.4}
    \begin{tabularx}{\textwidth}{|>{\hsize=.15\textwidth}X|X|}
    \hline
        \multicolumn{1}{|c|}{\textbf{Metric}}  & \multicolumn{1}{c|}{\textbf{Papers that used}} \\ \hline
        Precision & \cite{Su2016, Bora2016, ragothaman2016unsupervised, Indrabayu2017, oliveira_multi-objective_2017, Win22017, Bora2017, Saha2017, Taha2017, Hu2017, Tsukada2017, Zhang32017, Makris2017, YadanarWin2017, Taneja2017, Tareef2017, Gautam2017, AnousouyaDevi2018, Saha22018, abhinaav_abnormality_2019, Boecking2019, zhao_automated_2019, Zhang2019, Zhang22019, Sanyal2019, Zhao22019, saha2019prior, allehaibi2019segmentation, Ke2019, Saha2019, Hussain2020, lu2019deep, Hussain20202, Deng2020, sabeena_improved_2020, pmid32308878, Su2020, Ke2020, yang_interacting_2020, Khamparia2020, wang_nucleus_2020} \\ \hline
        Recall or Sensitivity & \cite{GarciaGonzalez2016, Phoulady2016, Zhao2016, Su2016, neghina_automatic_2016, Jele2016, Banerjee2016, Bora2016, song2016segmenting, ragothaman2016unsupervised, Indrabayu2017, oliveira_multi-objective_2017, Win22017, Bora2017, Win2017, Momenzadeh2017, Saha2017, Taha2017, Zhang22017, Hu2017, Tsukada2017, Zhang32017, Makris2017, SAHA22017, YadanarWin2017, Taneja2017, Tareef2017, Gautam2017, sangworasil_automated_2018, Momenzadeh2018, AnousouyaDevi2018, Bhowmik2018, Xu2018, Saha22018, Kar2019, William22019, Sunny2019, abhinaav_abnormality_2019, Wan2019, Khan22019, Teramoto2019, Boecking2019, william_automated_2019, Sompawong2019, zhao_automated_2019, Wang2019, Zhang2019, William2019, Win2019, Arya2019, Zhang22019, Aljakouch2019, lin2019fine, Sanyal2019, Sanghvi2019, Zhao22019, saha2019prior, allehaibi2019segmentation, Ke2019, Saha2019, Hussain2020, lu2019deep, Xiang2020, Hussain20202, Deng2020, BaykalKablan2020, sabeena_improved_2020, elliott2020application, Martinez2020, pmid32308878, Gonzalez2020, Ke2020, yang_interacting_2020, Khamparia2020, wang_nucleus_2020, Wasswa2019, Somasundaram2020}	 \\ \hline
        NPV & \cite{Makris2017, Kar2019, Boecking2019, Sanyal2019} \\ \hline
        FPR &  \cite{neghina_automatic_2016, song2016segmenting, oliveira_multi-objective_2017, Makris2017, Tareef2017, sangworasil_automated_2018, Xu2018, Wan2019, Teramoto2019, william_automated_2019, William2019, Hussain2020}	 \\ \hline
        FNR &   \cite{Phoulady2016, neghina_automatic_2016, song2016segmenting, oliveira_multi-objective_2017, Makris2017, Tareef2017, sangworasil_automated_2018, Xu2018, Wan2019, william_automated_2019, William2019, Hussain2020, Su2020} \\ \hline
        Specificity & \cite{GarciaGonzalez2016, Zhao2016, neghina_automatic_2016, Jele2016, Banerjee2016, Bora2016, Bora2017, Momenzadeh2017, Zhang22017, Makris2017, SAHA22017, Taneja2017, chankong_automatic_2018, Momenzadeh2018, AnousouyaDevi2018, Bhowmik2018, Kar2019, William22019, Khan22019, Teramoto2019, Boecking2019, william_automated_2019, Sompawong2019, Wang2019, gv_automatic_2019, William2019, Win2019, Arya2019, Aljakouch2019, lin2019fine, Sanyal2019, Sanghvi2019, saha2019prior, allehaibi2019segmentation, Hussain2020, Xiang2020, jia_parametric_2020, wang_adaptive_2020, BaykalKablan2020, elliott2020application, Martinez2020, Gonzalez2020, Somasundaram2020}	\\ \hline
        Accuracy & \cite{Zhao2016, Su2016, dumripatanachod2016centralized, kashyap2016cervical, Arya2016, Jele2016, Bora2016, song2016segmenting, Baykal2017, Bora2017, Momenzadeh2017, Mufidah2017, Taha2017, Zhang22017, ramdhani_hierarchical_2017, Garud22017, Makris2017, SAHA22017, Krau2017, lin2017llc, faturrahman_multi_2017, Taneja2017, Tareef2017, Dey2017, prum_abnormal_2018, sangworasil_automated_2018, chankong_automatic_2018, sharma2018cervical, Momenzadeh2018, nguyen2018deep, NirmalJith2018, AnousouyaDevi2018, Kowal2018, Bhowmik2018, deMesquitaSJunior2018, Dimauro2018, acharya2018segmentation, plissiti2018sipakmed, Kowal22018, han2018suspected, mishra2018theoretical, Kar2019, Khan2019, William22019, Sunny2019, abhinaav_abnormality_2019, Khan22019, Teramoto2019, william_automated_2019, Kilic2019, Sompawong2019, Wang2019, gv_automatic_2019, harangi_cell_2019, William2019, Win2019, Arya2019, Saikia2019, Solar2019, Zhang22019, Shanthi2019, Mousser2019, Arajo2019, Du2019, Aljakouch2019, lin2019fine, fekri-ershad_pap_2019, saha2019prior, allehaibi2019segmentation, Baykal2019, Hussain2020, lu2019deep, Xiang2020, jia_parametric_2020, Hussain20202, wang_adaptive_2020, Deng2020, BaykalKablan2020, jaya_channel_2020, Martinez2020, pmid32308878, Ke2020, Khamparia2020, gupta2020region, Nawandhar2020, Matias2020, Wasswa2019, Somasundaram2020}	 \\ \hline
        AUC &   \cite{Phoulady2017, Zhang22017, Hu2017, Bhowmik2018, deMesquitaSJunior2018, abhinaav_abnormality_2019, lin2019fine, Sanghvi2019, saha2019prior, Hussain2020, elliott2020application, Su2020, Gonzalez2020}	 \\ \hline
        MCC &  \cite{harangi_cell_2019, pmid32308878} \\ \hline
        Kappa Coefficient &   \cite{Taneja2017, Momenzadeh2018} \\ \hline
        Average Precision &   \cite{Hu2017, Zhang22019, Xiang2020}  \\ \hline
        DSC &   \cite{GarciaGonzalez2016, Phoulady2016, Su2016, Bora2016, song2016segmenting, bhan2016supervised, ragothaman2016unsupervised, Indrabayu2017, Phoulady2017, oliveira_multi-objective_2017, Win22017, Baykal2017, Bora2017, Saha2017, Zhang22017, Tsukada2017, Zhang32017, SAHA22017, YadanarWin2017, Garud2017, Taneja2017, Tareef2017, Gautam2017, Saha2018, Song2018, Bhowmik2018, Xu2018, Saha22018, abhinaav_abnormality_2019, Wan2019, Khan22019, Sompawong2019, zhao_automated_2019, gv_automatic_2019, Zhang2019, Das2019, Zhang22019, zhou2019irnet, Zhao22019, saha2019prior, allehaibi2019segmentation, Ke2019, Saha2019, lu2019deep, Hussain20202, BaykalKablan2020, sabeena_improved_2020, pmid32308878, yang_interacting_2020, Khamparia2020, lauw_mird-net_2020, wang_nucleus_2020, AlemiKoohbanani2020, AtkinsonAmorim2020, Matias2020, Wasswa2019}	  \\ \hline
        IoU &   \cite{SAHA22017, Taneja2017, Tareef2017, Bhowmik2018, harangi_cell_2019, zhou2019irnet, BaykalKablan2020, sabeena_improved_2020, bandyopadhyay2020segmentation, AtkinsonAmorim2020, Matias2020} \\ \hline
    \end{tabularx}
    \egroup
    \caption{Metrics citations}
\end{table}

\newpage

\bibliographystyle{elsarticle-num-names}
\bibliography{main.bbl}

\end{document}